\documentclass{article}

\usepackage{microtype}
\usepackage{graphicx}
\usepackage{subcaption}
\usepackage{placeins}
\usepackage{booktabs} 

\usepackage{hyperref}

\usepackage[accepted]{icml2026}

\usepackage{amsmath}
\usepackage{amssymb}
\usepackage{mathtools}
\usepackage{amsthm}
\usepackage{xspace}
\usepackage{multirow}
\usepackage{enumitem}
\usepackage{natbib}

\usepackage{colortbl}
\usepackage{xcolor}

\definecolor{ourslight}{gray}{0.96}
\definecolor{oursdark}{gray}{0.92}

\usepackage[capitalize,noabbrev]{cleveref}

\theoremstyle{plain}

\theoremstyle{definition}

\theoremstyle{remark}

\newcommand{\ours}{FRIGID\xspace}

\usepackage[textsize=tiny]{todonotes}

\icmltitlerunning{FRIGID: Scaling Diffusion-Based Molecular Generation from Mass Spectra at Training and Inference Time}

\begin{document}

\twocolumn[
    \icmltitle{FRIGID: Scaling Diffusion-Based Molecular Generation \\from Mass Spectra at Training and Inference Time}


  \icmlsetsymbol{equal}{*}

  \begin{icmlauthorlist}
    \icmlauthor{Montgomery Bohde}{equal,MIT,TAMU}
    \icmlauthor{Hongxuan Liu}{equal,MIT}
    \icmlauthor{Mrunali Manjrekar}{MIT}
    \icmlauthor{Magdalena Lederbauer}{MIT}
    \icmlauthor{Shuiwang Ji}{TAMU}
    \icmlauthor{Runzhong Wang}{MIT}
    \icmlauthor{Connor W. Coley}{MIT}
  \end{icmlauthorlist}

    \icmlaffiliation{MIT}{Massachusetts Institute of Technology, Cambridge, MA, United States}
    \icmlaffiliation{TAMU}{Texas A\&M University, College Station, TX, United States}

  \icmlcorrespondingauthor{Runzhong Wang}{runzhong@mit.edu}
  \icmlcorrespondingauthor{Connor W. Coley}{ccoley@mit.edu}

  \icmlkeywords{Machine Learning, ICML}

  \vskip 0.3in
]



\printAffiliationsAndNotice{\icmlEqualContribution}  

\begin{abstract}
Tandem mass spectrometry is prominent in scientific discovery workflows for identifying unknown small molecules, yet high-throughput structural elucidation remains challenging. While recent autoregressive and graph diffusion models have shown promise in \textit{de novo} elucidation, performance remains limited by poor scalability during both training and inference time. 
In this work, we present FRIGID, a framework with a novel diffusion language model that generates molecular structures conditioned on mass spectra via intermediate fingerprint representations and determined chemical formulae, training at the scale of hundreds of millions of unlabeled structures. We then demonstrate how forward fragmentation models enable inference-time scaling by identifying spectrum-inconsistent fragments and refining them through targeted remasking and denoising. While FRIGID already achieves strong performance with its diffusion base, inference-time scaling further improves its accuracy. FRIGID almost triples the previous state-of-the-art Top-1 accuracy on MassSpecGym and yields a 69.8\% relative improvement on NPLIB1, reaching 9.05\% and 14.16\%, respectively. Further empirical analyses show that FRIGID improves identification accuracy with additional inference-time compute, opening a promising new direction for continued improvements in \textit{de novo} structural elucidation. FRIGID code is publicly available at \url{https://github.com/coleygroup/FRIGID}.
\end{abstract}

\section{Introduction}

Tandem mass spectrometry (MS/MS) is a cornerstone of modern analytical chemistry, enabling large-scale profiling of small molecules in human health~\citep{gentry2024reverse}, drug discovery~\citep{prudent2021exploring}, and environmental chemistry~\citep{tian2021ubiquitous}, yet determining the chemical structure from an observed spectrum remains difficult. Most deployed identification pipelines are retrieval-based, matching a query spectrum against libraries of experimental or simulated spectra, or searching a structure database using predicted molecular fingerprints. 
Among them, retrieval-based spectrum simulators have demonstrated strong empirical accuracy~\citep{wang2025iceberg} and gained wide adoption~\citep{wang2021cfm-id4}.
However, retrieval pipelines fundamentally inherit the coverage limits of the underlying chemical libraries (e.g., PubChem) and cannot propose novel structures unless combined with an orthogonal method for molecular generation or enumeration \citep{Qiang2026-pa}. 

\begin{figure}[t!]
    \centering
    \includegraphics[width=0.94\linewidth]{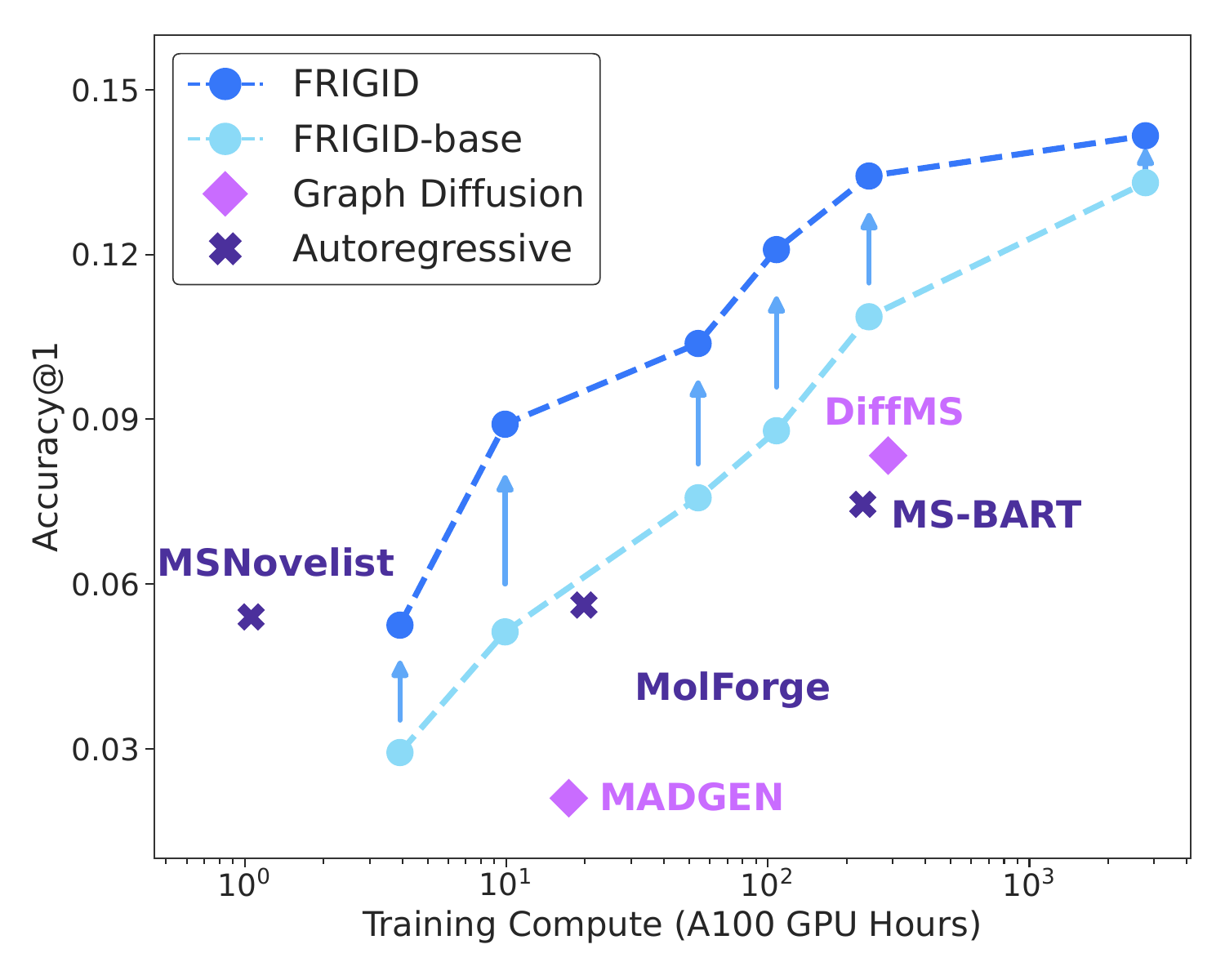}
    \vspace{-5pt}
    \caption{Top-1 exact-match accuracy as a function of training compute on the NPLIB1 dataset. Our \ours-base exhibits strong scaling behavior, further amplified by 10 rounds of inference-time refinement. Data points correspond to models trained on increasingly large corpora of unlabeled molecules.} 
    \label{fig:training_compute}
    \vspace{-15pt}
\end{figure}

\textit{De novo} elucidation models address this limitation by generating molecular structures directly from spectra. With the pioneering development of end-to-end models~\citep{shrivastava2021massgenie,butler2023ms2mol,litsa2023spec2mol}, two-stage architectures have demonstrated promising empirical performance. These methods typically consist of an encoder that first translates spectra into molecular fingerprints~\citep{duhrkop2015csifingerid,goldman2023mist}, followed by a decoder that translates fingerprints into molecular structures~\citep{le2020neuraldecipher,stravs2022msnovelist, ucak2023reconstruction,bohde2025diffms}. 
Fingerprint-to-molecule decoding is a critical step with ample room for improvement. Unlike molecular generation tasks in drug discovery campaigns, the core challenge of exact reconstruction in elucidation requires strong conditioning, self-correction, and robustness to fingerprint prediction errors.


Existing fingerprint-to-molecule decoders largely repurpose architectures originally developed for \textit{de novo} drug discovery, leaving the distinctive challenges and structure of MS/MS data insufficiently addressed. 
Autoregressive models~\citep{stravs2022msnovelist,han2025msbart,neo2025onesmallstep} condition on predicted fingerprints to generate SMILES sequences, whereas fragment-based representations---such as SAFE~\citep{noutahi2023safe}---remain comparatively underexplored despite their natural alignment with the fragment-centric nature of both fingerprints and MS/MS spectra. 
Discrete graph diffusion decoders~\citep{bohde2025diffms,wang2025madgenmassspecattendsnovo} introduce stronger inductive biases, but suffer from empirically slow inference.
Although fingerprint-structure pairs can be synthesized at virtually unlimited scale, enabling large-scale training as a promising path forward, systematic exploration of scaling effects remains limited to preliminary evidence in \citet{bohde2025diffms}. 
Finally, existing \textit{de novo} generators lack mechanisms for automatic correction of mistaken subgraphs, a capability that is crucial in the MS/MS context because a single token error typically leads to a wrong structural prediction.


To overcome these limitations, we introduce a diffusion language model that jointly enables scalable training, efficient inference, fragment-based generation, and inference-time error correction. As shown in Fig.~\ref{fig:training_compute}, our architecture exhibits favorable scaling behavior with increasing training data. In addition, we propose an inference-time scaling strategy that automatically corrects erroneous substructures---a capability absent from all existing \textit{de novo} generation methods. 
These design choices are motivated by the limitations of prior backbones: inference-time correction is nontrivial for standard autoregressive models, and graph diffusion models incur prohibitively slow inference (Table~\ref{table:diffusion_speed}).
Moving beyond naive inference-time scaling approaches such as longer denoising trajectories and stronger guidance~\citep{karras2022elucidating, ma2025inferencetimescalingdiffusionmodels}, we leverage forward fragmentation models, including ICEBERG~\citep{wang2025iceberg}, which attribute predicted spectral peaks to specific molecular fragments. This spectral consistency signal enables targeted masking and re-denoising of inconsistent subgraphs, rather than \textit{post hoc} filtering of generated candidates.

\textbf{Our contributions are summarized as follows:}

\begin{figure*}[t!]
  \centering
  \includegraphics[width=\textwidth]{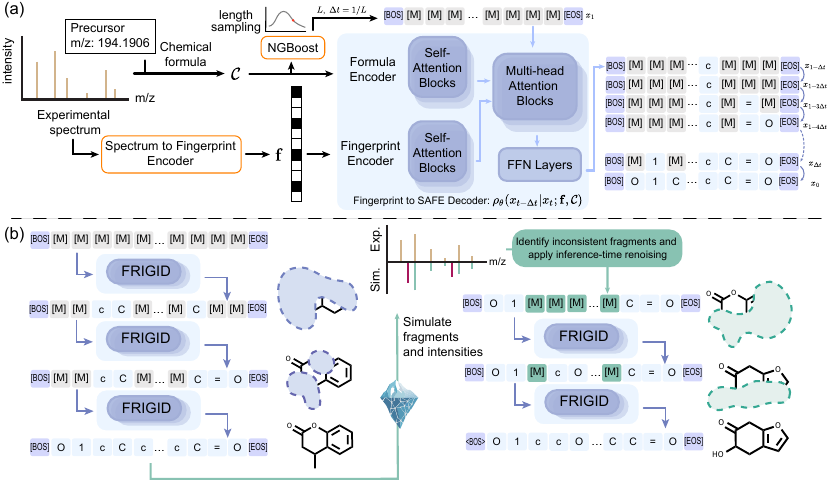}
  \vspace{-8pt}
  \caption{\textbf{(a)} FRIGID-base includes a masked diffusion language model (MDLM) that generates SAFE~\citep{noutahi2023safe} sequences. The MDLM decoder is conditioned on a precursor formula ($\mathcal{C}$) and a MIST~\citep{goldman2023mist}-predicted molecular fingerprint ($\mathbf{f}$), injected via cross-attention. SAFE sequence lengths are predicted using NGBoost~\citep{duan2020ngboost}. Generated structures are first filtered by chemical formula match, then ranked by Tanimoto similarity to the intermediate fingerprint $\mathbf{f}$. \textbf{(b)} The inference-time scaling pipeline enforces spectrum-structure consistency. ICEBERG model~\citep{wang2025iceberg} predicts simulated spectra for all generated structures. Discrepancies between the simulated and experimental spectra (hallucinations) are used to mask inconsistent tokens, which are then refined via targeted re-denoising.}
    \vspace{-10pt}
  \label{fig:contrib_overview}
\end{figure*}

\textbf{(1)} We present \ours, \textbf{\underline{F}}ragment \textbf{\underline{R}}efinement via \textbf{\underline{I}}CEBERG-\textbf{\underline{G}}uided \textbf{\underline{I}}nference \textbf{\underline{D}}iffusion, the first diffusion language model (DLM) framework for \textit{de novo} structural elucidation (Fig.~\ref{fig:contrib_overview}a). Our DLM backbone, \ours-base, generates 1D SAFE sequences conditioned on chemical formulae and utilizes MIST to translate spectra into fingerprints~\citep{goldman2023mist}. By adopting a language-based architecture, \ours-base can be readily pretrained on structure-only libraries encompassing hundreds of millions of structures. To further constrain the search space by molecular formula, a unique aspect of the MS/MS context, we use NGBoost~\citep{duan2020ngboost} to model the distribution of sequence lengths for a given molecular formula. In experiments, we demonstrate that \ours-base achieves strong performance even without any inference-time scaling and is significantly faster than peer diffusion models.
    
\textbf{(2)} We then demonstrate how forward fragmentation models such as ICEBERG can be leveraged for spectrum-structure cycle consistency by simulating spectra of generated molecules, identifying hallucinated fragments (i.e., mistaken subgraphs) that lack correspondence to experimental signals, and performing targeted remasking of subgraphs responsible for these inconsistencies (Fig.~\ref{fig:contrib_overview}b). This enables inference-time scaling that prioritizes inconsistent subgraphs.
    
\textbf{(3)} FRIGID achieves significant performance improvements over both diffusion and autoregressive baselines on established \textit{de novo} generation benchmarks. On the challenging MassSpecGym benchmark~\citep{bushuiev2024massspecgymbenchmarkdiscoveryidentification}, FRIGID almost triples the previous state-of-the-art Top-1 exact-match accuracy, reaching 9.05\%. On NPLIB1~\citep{duhrkop2021canopus}, FRIGID reaches 14.16\% Top-1 accuracy, a 69.8\% relative improvement over the previous state of the art, and achieves improved structural similarity even in cases where the true molecule is not recovered. 
    This work represents foundational progress towards the long-standing goal of computational structural elucidation from mass spectra.

\section{Related Work}

\textbf{Generative molecular design}.
Molecular representation is an important design factor for modern molecular generators. One family of models chooses linearized representations, which are convenient and fit soundly with infrastructures developed for natural language. SMILES~\citep{weininger1988smiles} enables the use of sequence models~\citep{gomez2018automatic,segler2018generating}.
Fragment-based alternatives, such as SAFE~\citep{noutahi2023safe}, support constrained editing and more stable generation in large chemical spaces. Efforts to approach \textit{de novo} structural elucidation using these linearized representations~\citep{stravs2022msnovelist,litsa2023spec2mol,butler2023ms2mol,shrivastava2021massgenie} have primarily been autoregressive in nature. 

Diffusion models~\citep{pmlr-v37-sohl-dickstein15,ho2020denoisingdiffusionprobabilisticmodels} have become a dominant paradigm for generative modeling and have been adapted to discrete and graph-structured domains~\citep{austin2023structureddenoisingdiffusionmodels,vignac2023digress}, demonstrating beneficial utility for molecular machine learning. These methods are particularly well-suited for underspecified conditional generation, where many valid outputs exist for a single condition---a common setting in chemical design. Theoretically, these models enable inference-time scaling through iterative renoising and denoising~\citep{li2023t2t,li2025generation}; however, it is computationally burdensome for graph data structures compared to linearized counterparts (e.g., SMILES).

Diffusion language models (DLMs)~\citep{li2022diffusion,austin2023structureddenoisingdiffusionmodels,sahoo2024mdlm} offer a balanced trade-off in efficiency while preserving the flexibility for inference-time scaling. An exemplary DLM, GenMol~\citep{lee2025genmol}, demonstrates strong capabilities in 
\textit{unconstrained} molecular generation of SAFE sequences in the context of drug discovery. 
These advancements establish the foundation for the core architecture of \ours-base.



\textbf{Structural elucidation from MS/MS}.
Most existing MS/MS identification pipelines remain retrieval-based. Classical methods search either experimental spectral libraries or larger structure databases using tandem MS, often by computing fragmentation trees to explain fragment relationships~\citep{bocker2016fragmentation}, predicting molecular fingerprints from spectra as in CSI:FingerID~\citep{duhrkop2015csifingerid}, or simulating spectra from candidate structures as in CFM-ID and later neural predictors~\citep{wang2021cfm-id4,young2024massformer,wang2025iceberg}. These approaches are highly effective when the true molecule or a close analog is represented among the candidates, but they cannot directly recover structures absent from the reference set.

Intermediate prediction modules have therefore become central building blocks of modern elucidation pipelines. Molecular formula inference methods, including fragmentation-tree-based approaches and learned models such as MIST-CF and BUDDY~\citep{goldman2023mist-cf,Xing2023-buddy}, help constrain the candidate search space. Fingerprint prediction from spectra, originally popularized by CSI:FingerID~\citep{duhrkop2015csifingerid} and more recently advanced by transformer-based spectrum encoders such as MIST~\citep{goldman2023mist}, converts MS/MS into structural descriptors that are straightforward to match against chemical databases or pass to downstream generators.

To move beyond retrieval, \textit{de novo} generators attempt to construct molecular structures from spectra themselves. Existing spectrum-to-molecule models span end-to-end approaches, such as MassGenie, Spec2Mol, and MS2Mol, that directly map spectra to molecular strings~\citep{shrivastava2021massgenie,litsa2023spec2mol,butler2023ms2mol}, as well as the now-dominant two-stage spectra-to-fingerprint-to-structure paradigm built around learned spectrum encoders such as MIST~\citep{goldman2023mist}. Representative examples include MSNovelist, which pairs CSI:FingerID with an autoregressive SMILES decoder~\citep{stravs2022msnovelist}, pipelines using general fingerprint decoders such as Neuraldecipher or MolForge~\citep{le2020neuraldecipher,ucak2023reconstruction,neo2025onesmallstep}, graph-based diffusion systems such as DiffMS~\citep{bohde2025diffms}, scaffold-guided generators such as MADGEN~\citep{wang2025madgenmassspecattendsnovo}, and unified sequence models such as MS-BART~\citep{han2025msbart}. In parallel, forward structure-to-spectrum models such as ICEBERG provide strong \textit{in silico} spectral prediction for candidate reranking~\citep{wang2025iceberg}. FOAM~\citep{manjrekar2026foam} further shows that forward fragmentation information can benefit inverse \textit{de novo} generation. \ours combines these two lines: it adopts the modern \textit{de novo} spectra-to-fingerprint-to-structure paradigm, while also leveraging the forward-modeling capability of ICEBERG \emph{inside} the generative loop for targeted iterative refinement rather than only for \textit{post hoc} scoring.

\section{\ours-Base Diffusion Model}
\label{sec:diffusion_backbone}

\subsection{Diffusion language generation}
As shown in Fig.~\ref{fig:contrib_overview}, we cast \textit{de novo} structural elucidation from MS/MS as a discrete conditional generation problem. Given a mass spectrum $\mathcal{S}$, \ours-base models a conditional distribution over SAFE strings~\citep{noutahi2023safe},
\begin{equation}
    p_\theta(\mathbf{x}_0\mid \mathcal{S}), \qquad \mathbf{x}_0=(x_{0,1},\ldots,x_{0,L})\in\mathcal{V}^L,
\end{equation}
where $\mathbf{x}_0$ is a SAFE string tokenized by byte-pair encoding (BPE), decomposing the string into frequently occurring substructure fragments, and $\mathcal{V}$ is the resulting fixed vocabulary containing additional special tokens: $\texttt{[BOS]}$, $\texttt{[EOS]}$, $\texttt{[MASK]}$, and $\texttt{[PAD]}$.
In practice, we turn $\mathcal{S}$ into compact conditioning signals: a chemical formula $\mathcal{C}$ inferred from MS1 and precursor information, and a spectrum-derived fingerprint $\mathbf{f}$ predicted by MIST~\citep{goldman2023mist} (see details in Sec.~\ref{sec:backbone_conditions}).
Instead of learning $p_\theta(\mathbf{x}_0\mid\mathcal{S})$ directly, our network learns $p_\theta(\mathbf{x}_0\mid \mathbf{f}, \mathcal{C})$.

\textbf{Masked diffusion language modeling.}
To parameterize $p_\theta(\mathbf{x}_0\mid \mathbf{f},\mathcal{C})$ non-autoregressively, we adopt a BERT-style masked diffusion language model (MDLM) following \citet{devlin2019bert,lee2025genmol}, as the denoiser over SAFE tokens. Intuitively, MDLM defines a forward corruption process that gradually converts each token into a $\texttt{[MASK]}$, and learns a reverse denoiser that imputes masked positions conditioned on the unmasked context and, in our case, the spectrum-derived conditions as well. 

Let $\mathbf{x}_0$ denote a clean SAFE token sequence. For a noise level $t\in[0,1]$, the forward process produces $\mathbf{x}_t$ by independently masking tokens with probability $1-\alpha(t)$:
\begin{equation}
    \begin{aligned}
    q_t(\mathbf{x}_t\mid \mathbf{x}_0)
    &= \prod_{i=1}^{L} q_t(x_{t,i}\mid x_{0,i}),\\
    q_t(x_{t,i}\mid x_{0,i})
    &=
    \begin{cases}
    \alpha(t), & x_{t,i}=x_{0,i},\\
    1-\alpha(t), & x_{t,i}=\texttt{[MASK]},\\
    0, & \text{otherwise},
    \end{cases}
    \end{aligned}
    \label{eq:forward_mask}
\end{equation}
where $\alpha(t)
= \exp\!\big((1 - t)\log \alpha_{\max} + t \log \alpha_{\min}\big)$ is a log-linear exponential noise schedule with $\alpha_{\max}=1$ and $\alpha_{\min}=10^{-3}$. The reverse model is a conditional denoiser that predicts the original tokens:
\begin{equation}
    p_\theta(\mathbf{x}_0 \mid \mathbf{x}_t, \mathbf{f},\mathcal{C})
    \;=\;
    \prod_{i=1}^{L} p_\theta(x_{0,i}\mid \mathbf{x}_t, \mathbf{f},\mathcal{C}),
    \label{eq:reverse_factorized}
\end{equation}
implemented as a Transformer encoder that outputs per-token categorical logits over $\mathcal{V}$. Our network has $N_{\text{layers}}{=}12$ Transformer blocks, hidden size $d{=}896$, and $H{=}14$ attention heads; the maximum token length is set to 256.

\textbf{Training objective.}
Following MDLM~\citep{sahoo2024mdlm}, we train by sampling $t\sim U(0,1)$ and minimizing the masked-token negative log-likelihood (ignoring $\texttt{[PAD]}$):
\begin{equation}
    \begin{aligned}
    \mathcal{L}_\text{MDLM}(\theta)
    &=
    \mathbb{E}_{\mathbf{x}_0\sim \mathcal{D},\,t\sim U(0,1)}
    \mathbb{E}_{\mathbf{x}_t\sim q_t(\cdot\mid \mathbf{x}_0)}\\
    &\cdot
    \Bigg[
    -\sum_{i: x_{t,i}=\texttt{[MASK]}}
    \log p_\theta(x_{0,i}\mid \mathbf{x}_t, \mathbf{f},\mathcal{C})
    \Bigg].
    \end{aligned}
    \label{eq:mdlm_loss}
\end{equation}
In practice, we use continuous-time training with antithetic time sampling for variance reduction and an exponential moving average (EMA) of model parameters for stability.

\textbf{Inference.}
To sample from the model, we first predict the target length $L$ (details for length prediction are in Sec.~\ref{sec:length_model}) and initialize $\mathbf{x}_1$ from a masked prior.
For each step $t$, the conditional denoiser predicts logits $\mathbf{z}_{t,i}\in\mathbb{R}^{|\mathcal{V}|}$ for each $i$ in masked positions $\mathcal{M}_t=\{i : x_{t,i}=\texttt{[MASK]}\}$, inducing a categorical distribution $\boldsymbol{\pi}_{t,i}=\mathrm{softmax}(\mathbf{z}_{t,i})$ with components $\pi_{t,i,v}$ for each $v\in\mathcal{V}$. We draw provisional tokens $\tilde{x}_{t,i}\sim\mathrm{Cat}(\boldsymbol{\pi}_{t,i})$ and compute a confidence score
$p_{t,i}=\max_{v\in\mathcal{V}} \pi_{t,i,v}$ (subsequently perturbed with annealed Gumbel noise to yield $\hat{p}_{t,i}$), then unmask the single \emph{most confident position} $i^{*}=\arg\max_{i\in \mathcal{M}_t} \hat{p}_{t,i}$ while keeping all other positions masked~\citep{sahoo2024mdlm,lee2025genmol}.
After $L$ steps, we decode the resulting SAFE sequence and deterministically convert it to SMILES, from which we generate RDKit objects for evaluation. Inference-time scaling design choices are discussed in Sec.~\ref{sec:inference_scaling}.

\subsection{Model architecture}
\label{sec:model_arch}

Our \ours-base architecture couples (i) an MS/MS encoder that produces conditioning fingerprint signals and (ii) a diffusion language backbone that decodes SAFE sequences under these conditions.

\textbf{Conditioning signals from MS/MS.}
\label{sec:backbone_conditions}
Given a spectrum $\mathcal{S}$, we use the MIST encoder~\citep{goldman2023mist} to predict $\mathbf{f}$, a 4096-bit radius-2 Morgan fingerprint~\citep{rogers2010ecfp}. For NPLIB1 and MassSpecGym, we train separate MIST encoders on experimental training spectra augmented with simulated spectra from 2 million PubChem molecules and the official 4 million MassSpecGym molecules, respectively. We use these dataset-specific checkpoints for both \ours and MolForge; training details are provided in Appendix~\ref{appendix:mist_training}. The chemical formula $\mathcal{C}$ can typically be inferred upstream from high-resolution MS1 measurements 
using dedicated formula annotation tools~\citep{bocker2016fragmentation,Xing2023-buddy,goldman2023mist-cf}. 



\textbf{Conditioner embedding and cross-attention.}
We encode and concatenate both $\mathcal{C}$ and $\mathbf{f}$ into a single conditioning \emph{embedding sequence} $\mathbf{Z} \triangleq [\mathbf{Z}^{(\mathcal{C})},\, \mathbf{Z}^{(\mathbf{f})}]$ and inject it via cross-attention in every multi-head Transformer block. For the formula, we represent $\mathcal{C}$ as a count vector $\mathbf{y}\in\mathbb{N}^{|\mathcal{E}|}$ over a fixed element set $\mathcal{E}$ ($|\mathcal{E}|{=}30$ in our implementation) and embed each element type together with its count, yielding a fixed-length formula embedding sequence $\mathbf{Z}^{(\mathcal{C})}\in\mathbb{R}^{|\mathcal{E}|\times d}$ and a padding mask for zero-count elements.

For the fingerprint $\mathbf{f}$, we identify a set of active bit indices by selecting MIST-predicted probabilities exceeding $\vartheta=0.2042523548$ on NPLIB1 and $\vartheta=0.1879101470$ on MassSpecGym\footnote{The threshold for each checkpoint is calibrated to match the total predicted on-bit count to the reference on-bit count on validation spectra, then fixed for test evaluation.}. We retain a maximum of 256 indices following the canonical bit order, and map each index to a learnable embedding. We then apply $3$ layers of self-attention without positional encodings to these embeddings, making the network permutation-equivariant while learning interactions among active substructures, yielding $\mathbf{Z}^{(\mathbf{f})}\in\mathbb{R}^{L_{\mathbf{f}}\times d}$ with $L_{\mathbf{f}}\le 256$ and a padding mask. In our cross-attention layer, as shown in Fig.~\ref{fig:contrib_overview}, we concatenate $\mathbf{Z}^{(\mathcal{C})}$ and $\mathbf{Z}^{(\mathbf{f})}$ along the conditioning-length dimension (and concatenate their padding masks accordingly), resulting in a total conditioning length $L_{\text{cond}} = |\mathcal{E}| + L_{\mathbf{f}}$.
Each cross-attention layer is
followed by an output projection and residual layer normalization~\citep{vaswani2017attention}. 

\textbf{Token length prediction.}
\label{sec:length_model}
The diffusion process requires a target length $L$ to initialize the masked sequence $\mathbf{x}_1$.
We learn a formula-to-length predictor $p_\phi(L\mid \mathcal{C})$ using NGBoost~\citep{duan2020ngboost}, which uses a boosted ensemble of regression trees with natural-gradient updates to predict the parameters of a conditional Normal distribution from an element-count feature vector $\mathbf{a}(\mathcal{C})\in\mathbb{R}^{|\mathcal{E}|}$. This is well-suited as a lightweight model for the length prediction task, where $L$ is inherently uncertain (multiple structures can share $\mathcal{C}$ but yield different tokenization lengths).
At inference, we obtain $(\mu_\phi(\mathcal{C}),\sigma_\phi(\mathcal{C}))$ and sample
$
    \tilde{L} \sim \mathcal{N}\!\big(\mu_\phi(\mathcal{C}),\, \lambda\,\sigma_\phi^2(\mathcal{C})\big), \
    L = \mathrm{round}(\tilde{L}),
$
clipped to a valid range, and create a fully-masked $\mathbf{x}_1=[\texttt{[BOS]}, \texttt{[MASK]},\ldots,\texttt{[MASK]}, \texttt{[EOS]}]$ with $L{+}2$ tokens (including $\texttt{[BOS]}$/$\texttt{[EOS]}$), before padding to the batch maximum. The scale factor $\lambda$ controls the sampling variance around $\mu_\phi(\mathcal{C})$, trading off exploration of plausible lengths against wasted compute on implausible lengths.

\section{Cycle-Consistent Inference-Time Scaling}
\label{sec:inference_scaling}

\begin{figure*}[t!]
  \centering
  \includegraphics[width=0.95\textwidth]{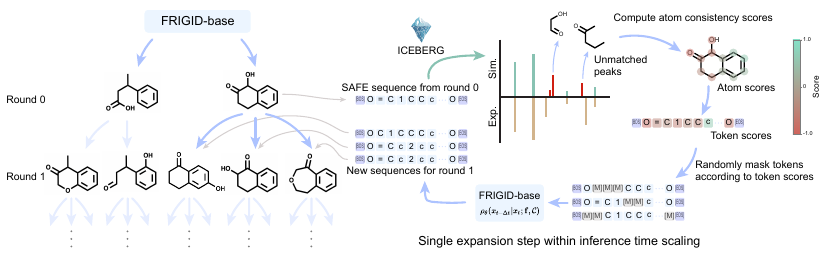}
  \caption{Mechanism of ICEBERG-guided inference-time scaling. \textbf{Left}: the process first comprises a beam search, starting with structures generated by \ours-base (Round 0). \textbf{Right}: at each expansion step, we use ICEBERG to simulate a mass spectrum, compare it against the experimental spectrum to identify hallucinated peaks---peaks present in simulation but missing in the experiment. Atom-wise consistency scores ($S_a$) are collected based on an atom's presence in fragments corresponding to matched or hallucinated peaks (Eq.~(\ref{eq:atom_scores})). These scores are aggregated at the token-level to define a masking probability $P_\text{mask}$ (Eq.~(\ref{eq:p_mask})), to remask inconsistent substructures. The masked sequence is then re-denoised by \ours-base to generate refined candidates, constructing candidates for Round 1.}
  \label{fig:inference_time_overview}
    \vspace{-10pt}
\end{figure*}


While our diffusion decoder provides a strong prior over valid chemical structures, the model may still generate structures that are inconsistent with the observed spectrum. To mitigate this, we introduce an iterative refinement strategy for spectrum-structure cycle consistency (Fig.~\ref{fig:inference_time_overview}).

\subsection{Forward fragmentation simulation}
ICEBERG~\citep{wang2025iceberg} is a state-of-the-art forward fragmentation model that predicts tandem mass spectra from molecular structures, working in the opposite direction of \ours. 
Driving its strong empirical accuracy in spectral prediction, 
ICEBERG simulates spectra by predicting the most likely fragment substructures of the candidate structure and scores each fragment with a predicted intensity. 
Formally, an ICEBERG simulated spectrum is a set of $k$ tuples $\mathcal{S}_{\text{sim}} = \{(\mathcal{G}_j, m_j, I_j)\}_{j=1}^k$, one tuple per peak in the simulated spectrum. For each peak $j$, $\mathcal{G}_j$ is a fragment (subgraph) of the original molecule, $m_j$ is the mass-to-charge ratio of the fragment $\mathcal{G}_j$, and $I_j$ is the predicted intensity. 
Our core insight is that we can use ICEBERG's predicted fragments to determine which atoms in a generated structure are incompatible with the observed spectrum.

\begin{algorithm}[tb!]
\caption{
ICEBERG-Guided Inference-Time Scaling}
\label{alg:refinement}
\begin{algorithmic}
\STATE {\bfseries Input:} Observed Spectrum $\mathcal{S}_{\text{obs}}$, MIST-Predicted Fingerprint $\mathbf{f}$, Chemical Formula $\mathcal{C}$, Rounds $R$, Batch size $B$, Top-$M$ count $M$, Number of renoised samples $N$
\STATE $\mathcal{X}_{\text{pool}} \leftarrow \emptyset$
\FOR{$r=1$ {\bfseries to} $R$}
    \IF{$r=1$}
        \STATE Sample lengths $L_i \sim \mathcal{N}(\mu_\phi(\mathcal{C}), \lambda \sigma_\phi^2(\mathcal{C}))$
        \STATE $\mathcal{X}_{\text{pool}} \leftarrow$ Generate $B$ samples $\mathcal{X}_{\text{new}}$ using FRIGID-base with lengths $L_i$
    \ELSE
        \STATE $\mathcal{X}_{\text{top}} \leftarrow \underset{\mathbf{x} \in \mathcal{X}_{\text{pool}}}{\operatorname{arg\,max}_M} \; \text{TaniSim}(fp(\mathbf{x}), \mathbf{f})$
        \STATE $\mathcal{X}_{\text{new}} \leftarrow \emptyset$
        \FOR{$\mathbf{x} \in \mathcal{X}_{\text{top}}$}
            \STATE $\mathcal{S}_{\text{sim}} \leftarrow \text{ICEBERG}(\mathbf{x})$
            \STATE Compute atom consistency scores $S_a$ using $\mathcal{S}_{\text{sim}}$ and $\mathcal{S}_{\text{obs}}$ using Eq.~(\ref{eq:atom_scores})
            \STATE Calculate token masking probabilities $P_{\text{mask}}$ from $S_a$ using Eq.~(\ref{eq:p_mask})
            \STATE Construct $N$ renoised inputs $\tilde{\mathbf{x}}$ by masking $\mathbf{x}$ using $P_{\text{mask}}$
            \STATE $\mathcal{X}_{\text{new}} \leftarrow \mathcal{X}_{\text{new}} \cup \text{Denoise}(\tilde{\mathbf{x}} | \mathbf{f}, \mathcal{C})$
        \ENDFOR
    \ENDIF
    \STATE $\mathcal{X}_{\text{pool}} \leftarrow \mathcal{X}_{\text{pool}} \cup \mathcal{X}_{\text{new}}$
\ENDFOR
\STATE \textbf{return} $\mathcal{X}_{\text{pool}}$
\end{algorithmic}
\end{algorithm}

\subsection{ICEBERG-guided error refinement}

We run inference-time scaling for $R$ rounds. In the first round, we simply generate $B$ samples from the \ours-base with different sequence lengths $L_i\sim \mathcal{N}\!(\mu_\phi(\mathcal{C}),\, \lambda\,\sigma_\phi^2(\mathcal{C}))$. In all future rounds, we use ICEBERG for targeted remasking and denoising.

Let $\mathcal{X}_{r-1}$ be the set of all generated molecules in rounds $1, \dots, r-1$. In each round $r > 1$, we select the top $M$ molecules $\mathcal{X}_{\text{top}}$ ranked by Tanimoto similarity to the MIST-predicted fingerprint $\mathbf{f}$.
For each sample $\mathbf{x}_0^{(i)}\in \mathcal{X}_{\text{top}}$, we then use ICEBERG to simulate the spectrum $\mathcal{S}_{\text{sim}}$ and attempt to match the simulated peaks in $\mathcal{S}_{\text{sim}}$ to peaks in the observed spectrum $\mathcal{S}_{\text{obs}} = \{(m_{\text{obs}}, I_{\text{obs}})\}_{i=1}^l$ within a 20 ppm (parts-per-million) tolerance as $\mathcal{P}_{\text{match}}$, 
and (with slight abuse of notation) we label all unmatched peaks $\mathcal{P}_{\text{halluc}} = \mathcal{S_{\text{sim}}} \setminus \mathcal{P}_{\text{match}}$ as hallucinations.

We then compute a consistency score $S_a$ for every atom $a$ in sample $\mathbf{x}_0^{(i)}$. Let $\text{atoms}(\mathcal{G}_j)$ be the set of atoms involved in fragment $\mathcal{G}_j$. Atoms can be (and usually are) part of multiple fragments. We define the score for each atom $a$ by aggregating over the fragments $a$ is in:
\begin{equation}
    S_a = \sum_{\mathcal{G}_j\in{\mathcal{S}_{\text{sim}}}}\mathbb{I}\left[a \in \text{atoms}(\mathcal{G}_j)\right]\cdot \delta(\mathcal{G}_j),
    \label{eq:atom_scores}
\end{equation}
where $\delta(\mathcal{G}_j)$ is a reward function that assigns $+$1 if $\mathcal{G}_j\in \mathcal{P}_\text{match}$ and $-\gamma$ if $\mathcal{G}_j\in \mathcal{P}_\text{halluc}$. Intuitively, atoms that frequently participate in spectrum-consistent peaks accumulate positive scores while atoms in hallucinated fragments accumulate negative scores. The parameter $\gamma$ allows us to trade off between rewarding matched atoms and penalizing spectrum-inconsistent atoms. We then normalize the atom-wise consistency scores such that $\tilde{S}_a \in [-1, 1]$.

Since our \ours-base operates over SAFE strings and each token may cover multiple atoms (e.g. in a ring or functional group) or similarly connect them (e.g., bond order tokens), we must aggregate atom consistency scores into token-level consistency scores. Let $A_j$ be the set of atoms associated with token $x_{j}$. The score for token $x_{j}$ is determined by the worst atom it covers,
$
    S_{\text{token}}(j) =  \min_{a \in A_j} \tilde{S}_a.
$

This $\min$ aggregation ensures that if any part of the token is predicted to be responsible for a hallucinated fragment, it will still receive a negative score. Finally, we convert token scores into masking probabilities $P_{\text{mask}}(j)$ for each token:
\begin{equation}
\begin{split}
    P_{\text{mask}}(j) = 
    \begin{cases} 
        p_{\max} \cdot |S_{\text{token}}(j)| & \text{if } S_{\text{token}}(j) \le 0, \\
        0 & \text{if } S_{\text{token}}(j) > 0.
    \end{cases}
    \label{eq:p_mask}
\end{split}
\end{equation}

For each $\mathbf{x}_0^{(i)}\in \mathcal{X}_{\text{top}}$, we construct $N$ renoised inputs $\mathbf{x}_t^{(i)}$ for refinement by randomly and independently masking each token $x_j \in \mathbf{x}_0^{(i)}$ with probability $P_{\text{mask}}(j)$. These new samples are then denoised with our backbone and added to $\mathcal{X}_r$ to be ranked in the following round. The full \ours inference-time scaling approach is detailed in Algorithm~\ref{alg:refinement}.

\begin{table*}[t!]
\caption{\textit{De novo} structural elucidation performance on NPLIB1 \citep{duhrkop2021canopus} and MassSpecGym~\citep{bushuiev2024massspecgymbenchmarkdiscoveryidentification} datasets, assuming known chemical formulae. The best performing model for each metric is \textbf{bold} and the second best is \underline{underlined}.
$\dag$~indicates our implementations of baseline approaches without public code. Methods are approximately ordered by performance. \ours and MolForge use MIST 2M-aug on NPLIB1 and MIST 4M-aug on MassSpecGym. \ours uses 10 refinements on both datasets. For MS-BART on MassSpecGym, we report accuracy based on 2D InChIKey matching from Appendix~C of~\citet{han2025msbart}, rather than the full InChIKey matching accuracy reported in the main text.
}
\vspace{-0.05in}
\label{table:main}
\begin{center}
{
\begin{tabular}{lcccccc}
\toprule
& \multicolumn{3}{c}{Top-1} & \multicolumn{3}{c}{Top-10} \\
\cmidrule(lr){2-4}
\cmidrule(lr){5-7}
Model & Accuracy $\uparrow$ & MCES $\downarrow$ & Tanimoto $\uparrow$ & Accuracy $\uparrow$ & MCES $\downarrow$ & Tanimoto $\uparrow$ \\
\midrule
\midrule
& \multicolumn{4}{c}{NPLIB1} & & \\
\midrule
Spec2Mol & 0.00\% & 48.57 & 0.12 & 0.00\% & 17.58 & 0.16 \\
MADGEN & 2.10\% & 20.56 & 0.22 & 2.39\% & 12.69 & 0.27 \\
MIST + Neuraldecipher & 2.32\% & {10.26} & 0.35 & 6.11\% & {8.53} & 0.43 \\
MIST + MolForge$^\dag$ & 5.62\% & 20.72 & 0.43 & 11.36\% & 10.55 & 0.52 \\
MIST + MSNovelist & {5.40\%} & 13.10 & 0.34 & {11.04}\% & 8.57 & {0.44} \\
MS-BART & 7.45\% & 9.66 & 0.44 & 10.99\% & 8.31 & 0.51 \\
DiffMS & 8.34\% & 10.46 & 0.35 & 15.44\% & 8.54 & {0.47} \\
\midrule
\ours-base (ours) & \underline{13.31\%} & \underline{7.58} & \underline{0.50} & \underline{20.15\%} & \underline{6.16} & \underline{0.56} \\
\ours (ours) & \textbf{14.16\%} & \textbf{7.16} & \textbf{0.51} & \textbf{25.27\%} & \textbf{5.79} & \textbf{0.58} \\
\midrule
\midrule
& \multicolumn{4}{c}{MassSpecGym} & & \\
\midrule
MIST + MSNovelist & 0.00\% & 39.84 & 0.06 & 0.00\% & 18.83 & 0.15 \\
Spec2Mol & 0.00\% & 36.78 & 0.12 & 0.00\% & 36.02 & 0.16 \\
MIST + Neuraldecipher & 0.00\% & 22.93 & 0.14 & 0.00\% & 21.76 & 0.16 \\
MADGEN & {1.31\%} & 27.47 & {0.20} & {1.54}\% & {16.84} & {0.26} \\
MS-BART & 1.26\% & 16.47 & 0.23 & 1.28\% & 15.12 & 0.28 \\
DiffMS & 2.30\% & 13.96 & 0.28 & 4.25\% & 11.68 & {0.39} \\
FOAM (DiffMS + PubChem) & 1.50\% & \textbf{12.21} & 0.35 & \underline{10.28\%} & \textbf{6.06} & \textbf{0.53} \\
MIST + MolForge$^\dag$ & 3.29\% & 31.77 & 0.31 & 5.65\% & 16.67 & 0.40 \\
\midrule
\ours-base (ours) & \underline{8.21\%} & 13.16 & \underline{0.38} & 9.82\% & 11.65 & 0.43 \\
\ours (ours) & \textbf{9.05\%} & \underline{12.60} & \textbf{0.39} & \textbf{12.23\%} & \underline{11.41} & \underline{0.43} \\
\bottomrule
\end{tabular}
}
\end{center}
\vskip -0.1in
\end{table*}

\section{Experiments}

\subsection{Evaluation metrics}

We adopt the standard \textit{de novo} generation metrics of \citet{bushuiev2024massspecgymbenchmarkdiscoveryidentification} used by all recent works~\citep{wang2025madgenmassspecattendsnovo, bohde2025diffms, han2025msbart}:
\begin{itemize}[noitemsep,topsep=0pt]
    \item \textbf{Top-$k$ accuracy:} measures whether a molecule with the same connectivity-based 2D InChIKey as the true molecule is ranked in the Top-$k$ predictions
    \item \textbf{Top-$k$ maximum Tanimoto similarity:} measures the highest structural similarity to the true molecule among the Top-$k$ predictions
    \item \textbf{Top-$k$ minimum MCES:} measures the smallest graph edit distance (maximum common edge subgraph) to the true molecule among the Top-$k$ predictions
\end{itemize}
In Appendix~\ref{sec:addl_metrics}, we report additional close/meaningful match metrics established by~\citet{butler2023ms2mol}.

\subsection{Datasets and baselines}

\textbf{Datasets and benchmarks}. Following prior works
, we evaluate \ours on the NPLIB1~\citep{duhrkop2021canopus} and MassSpecGym~\citep{bushuiev2024massspecgymbenchmarkdiscoveryidentification} benchmarks. The MassSpecGym benchmark poses a challenging out-of-distribution shift with a scaffold-based test split that ensures no training molecules have an MCES $<10$ compared to any test molecules. While historically some works also evaluated on NIST datasets~\citep{nist_database}, since this reference library is not publicly available, we focus on these two open-source libraries to enable comparison to the most recent works. In line with peer methods, we focus on the setting where the ground-truth chemical formula is provided.

\textbf{\ours training}. While prior works typically train for many epochs on a smaller dataset such as the 2.8M unlabeled molecules used by DiffMS and MIST + MolForge, we draw from LLM pretraining and train for a single epoch on a large corpus up to hundreds of millions of unlabeled structures. 
Specifically, we train our \ours-base DLM on up to 1 billion structures from \citet{noutahi2023safe}, excluding NPLIB1 and MassSpecGym validation and test structures by connectivity-based (2D) InChIKey. Decoder training details are provided in Appendix~\ref{appendix:decoder_training}. We train the ICEBERG models used for inference-time scaling on each benchmark's experimental training split (Appendix~\ref{appendix:iceberg_training}).

\textbf{MIST training}. We train separate MIST encoders~\citep{goldman2023mist} from scratch on each benchmark's experimental training spectra, augmented with spectra simulated by GLACIER~\citep{wang2026glacier} from 2 million PubChem molecules for NPLIB1 (MIST 2M-aug) and the official 4 million unlabeled molecules for MassSpecGym (MIST 4M-aug). Both \ours and our MolForge reimplementation use these encoders. Training details are provided in Appendix~\ref{appendix:mist_training}. A previous version of this manuscript used synthetic data augmentation that, while excluding any exact overlap with the test structures, made inappropriate use of test structure information. These prior results are discussed in Appendix~\ref{appendix:mist_training}. 

\textbf{Peer methods}. Prior works fall into two categories: Spec2Mol~\citep{litsa2023spec2mol}, MIST + Neuraldecipher~\citep{le2020neuraldecipher, bohde2025diffms}, MIST + MolForge~\citep{neo2025onesmallstep}, MIST + MSNovelist~\citep{stravs2022msnovelist}, and MS-BART~\citep{han2025msbart} are all autoregressive models that predict SMILES/SELFIES sequences from spectra, while MADGEN~\citep{wang2025madgenmassspecattendsnovo}, DiffMS~\citep{bohde2025diffms}, and FOAM~\citep{manjrekar2026foam} are graph-based approaches, with FOAM further augmenting a forward-model-guided iterative optimization pipeline using DiffMS-generated + PubChem formula-matched seed candidates. Additionally, all baselines except for Spec2Mol and MADGEN rely on MIST~\citep{goldman2023mist} to extract dense structural features from spectra. 

Because \citet{neo2025onesmallstep} do not report results on NPLIB1 and do not have publicly available code, we re-implemented MIST + MolForge and report our reproduced results on both benchmarks. Appendix~\ref{appendix:molforge_discrepancy} describes the MIST checkpoint and inference settings used to evaluate MolForge. \citet{mismetti2026testtimetunedlanguagemodels} recently developed a test-time tuning strategy for \textit{de novo} structural elucidation; however, we exclude this method from comparison because it pretrains its model on a dataset~\citep{alberts2024unraveling} without confirmation that NPLIB1 and MassSpecGym test structures are excluded during training.
We provide more details on this exclusion in Appendix~\ref{appendix:leakage}. We also note that their test-time tuning is orthogonal to our inference-time scaling, and future works may benefit from combining both ideas.  

\subsection{Results and Discussion}

\textbf{Main results}. As seen in Table~\ref{table:main}, both \ours-base and \ours achieve leading performance across both benchmarks, achieving state-of-the-art results across 9 of the 12 metrics. On the more challenging MassSpecGym benchmark, \ours almost triples the previous state-of-the-art Top-1 accuracy, increasing it from 3.29\% to 9.05\%. On NPLIB1, \ours achieves 14.16\% Top-1 accuracy, a 69.8\% relative improvement over the leading baseline, and improves structural similarity even in cases where the true molecule could not be recovered. While \ours achieves 3.9$\times$ and 6.0$\times$ the exact-match accuracy of DiffMS and FOAM, respectively, these graph-based models still maintain competitive performance on the graph-based distance metric, MCES. In Appendix~\ref{sec:addl_metrics} and Appendix~\ref{appendix:scaling_ablations}--\ref{appendix:e2e_ablation}, we report additional close/meaningful-match metrics, inference-time scaling ablations, robustness analyses, and direct spectra-conditioning ablations. Our DLM achieves strong performance on both benchmarks by itself; however, our proposed inference-time scaling offers performance improvements across both benchmarks.

\textbf{Computational efficiency}.
\label{sec:ablate_speed}
Table~\ref{table:diffusion_speed} compares the inference runtime of the \ours backbone against other diffusion models. 
Despite using an architecture with 247.0M parameters, \ours achieves an inference speed of 6.58 seconds per spectrum, 20$\times$ faster than the graph-based DiffMS (131.2 s/spectrum) and over 3.6$\times$ faster than MADGEN (23.80 s/spectrum). This significant speedup can be attributed to the efficiency of our parallel decoding scheme compared to the computationally expensive graph operations required by existing diffusion models. This low latency provides the computational budget necessary to perform massive inference-time scaling (next paragraph) while remaining practical for large-scale spectral annotation.

\begin{table}[t!]
\caption{Number of parameters and inference throughput of \ours backbone compared with baseline diffusion models. Throughput evaluated using random samples from the NPLIB1 dataset on a single NVIDIA 6000 Ada GPU.}
\label{table:diffusion_speed}
\centering
\resizebox{\columnwidth}{!}
{
\begin{tabular}{l|ccc}
\toprule
\multirow{2}{*}{Model} & \multirow{2}{*}{\# Parameters} & Inference Latency  $\downarrow$ & Relative \\
&  & (sec. / spectrum) & Speedup \\
\midrule
DiffMS & 53.2M & 131.2 & 1.00$\times$ \\
MADGEN & 2.51M & 23.80 & 5.51$\times$ \\
\ours & 247.0M & 6.58 & 20.0$\times$ \\
\bottomrule
\end{tabular}}
\vspace{-10pt}
\end{table}

\begin{table*}[t!]
\caption{ICEBERG-guided refinement applied to DiffMS on the NPLIB1 benchmark. ICEBERG improves DiffMS substantially, confirming that the refinement strategy is model-agnostic; however, the resulting latency is much higher than for \ours.}
\label{table:diffms_iceberg}
\centering
{\small
\begin{tabular}{lccccccc}
\toprule
& \multicolumn{3}{c}{Top-1} & \multicolumn{3}{c}{Top-10} & Inference Compute \\
\cmidrule(lr){2-4}
\cmidrule(lr){5-7}
Model & Accuracy $\uparrow$ & MCES $\downarrow$ & Tanimoto $\uparrow$ & Accuracy $\uparrow$ & MCES $\downarrow$ & Tanimoto $\uparrow$ & (sec./sample) $\downarrow$ \\
\midrule
DiffMS & 8.34\% & 10.46 & 0.35 & 15.44\% & 8.54 & 0.47 & 131.21 \\
DiffMS + ICEBERG & 13.33\% & 9.30 & 0.43 & 23.29\% & 6.91 & 0.54 & 1423.30 \\
\midrule
\ours-base (ours) & 13.31\% & 7.58 & 0.50 & 20.15\% & 6.16 & 0.56 & 6.58 \\
\ours (ours) & 14.16\% & 7.16 & 0.51 & 25.27\% & 5.79 & 0.58 & 41.3 \\
\bottomrule
\end{tabular}}
\vspace{-2mm}
\end{table*}

\begin{figure}[t!]
    \centering
    \includegraphics[width=1.0\linewidth]{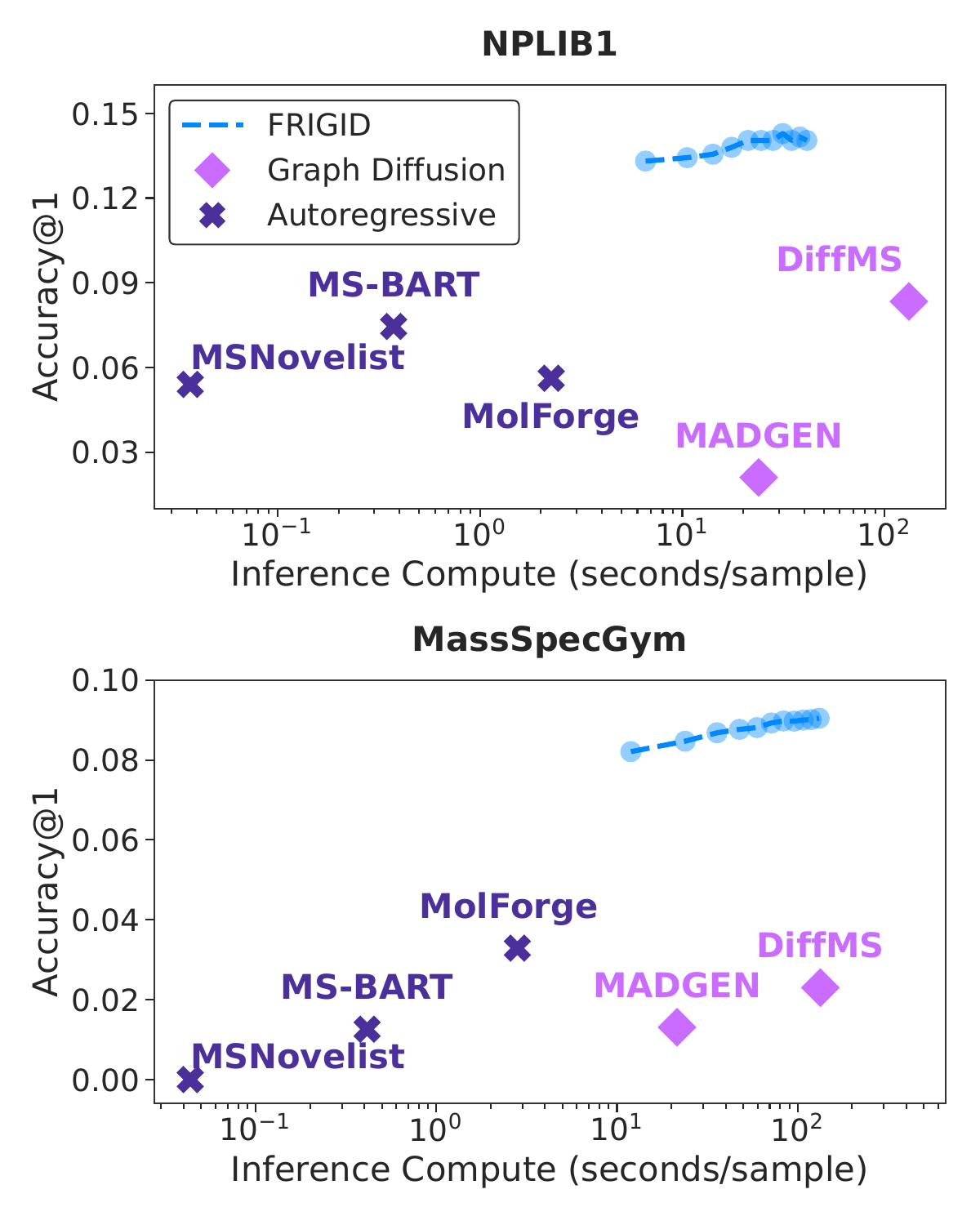}
    \caption{Inference-time scaling on NPLIB1 and MassSpecGym. The plots display Top-1 identification accuracy as a function of inference compute time for \ours and baseline methods. The blue dashed line illustrates \ours{}'s performance from initial generation through 10 rounds of refinement. \ours significantly outperforms autoregressive baselines (e.g., MS-BART, MSNovelist) and graph diffusion models (e.g., DiffMS, MADGEN) in accuracy, while remaining more efficient than graph diffusion models.
    } 
    \label{fig:inference_scaling}
    \vspace{-15pt}
\end{figure}

\textbf{Inference-time compute scaling}.
\label{sec:inference_scaling_results}
Fig.~\ref{fig:inference_scaling} highlights the trade-off between identification accuracy and inference cost. \ours improves Top-1 accuracy with additional refinement compute (blue dashed line). Ten refinements yield a 6.4--10.2\% relative improvement over \ours-base (Table~\ref{table:main}). Even the base model, without inference-time correction, outperforms both autoregressive and graph diffusion baselines in accuracy while remaining 20 times faster than graph-based methods like DiffMS. This establishes \ours as a flexible state-of-the-art that allows users to trade compute for precision on difficult samples.

To test whether ICEBERG-guided refinement is specific to \ours, we also apply the same refinement procedure to DiffMS. On NPLIB1, DiffMS improves from 8.34\% to 13.33\% Top-1 accuracy and from 0.35 to 0.43 Top-1 Tanimoto (Table~\ref{table:diffms_iceberg}), confirming that the refinement mechanism itself is model-agnostic. However, this improvement requires 1423.30 seconds per sample, compared to approximately 41.3 seconds for \ours. This reinforces our core claim: forward fragmentation guidance transfers across backbones, but a sequence-based diffusion decoder is what makes iterative refinement practical. Appendix~\ref{appendix:scaling_ablations} reports sensitivity analyses over refinement hyperparameters.

\textbf{Pretraining scaling analysis}.
\label{sec:ablate_train_scaling}
As illustrated in Figure~\ref{fig:training_compute}, our DLM backbone exhibits strong scaling behavior with respect to training compute. Top-1 accuracy on NPLIB1 increases with the amount of compute used during pretraining (approximated by training duration on increasingly large datasets). Notably, even at lower compute budgets, \ours pushes the accuracy significantly beyond existing baselines, achieving higher accuracy with less training cost than graph-based diffusion models like MADGEN. This trend suggests that our sequence-based architecture effectively leverages large-scale unlabeled data, and that further scaling of the pretraining corpus and model size is a promising avenue for continued performance gains.

\section{Conclusion}

In this work, we present \ours, a DLM-based framework for \textit{de novo} structural elucidation from tandem mass spectra.
In favor of a sequence-based diffusion backbone, we achieve a 20$\times$ inference speedup over state-of-the-art graph diffusion methods, enabling scalable decoder pretraining on hundreds of millions of molecules without spectrum labels. Our efficient DLM architecture enables novel inference-time scaling, leveraging a neural spectral simulator to iteratively detect and correct spectral hallucinations through targeted masking and denoising. Empirically, \ours almost triples the previous state-of-the-art Top-1 accuracy on MassSpecGym and yields a 69.8\% relative improvement on NPLIB1, reaching 9.05\% and 14.16\%, respectively. These results suggest a promising new direction for continued improvements in \textit{de novo} structural elucidation.





\section*{Software and Data}
FRIGID code and models are available at \url{https://github.com/coleygroup/FRIGID}.


\section*{Acknowledgments}

This work was partly supported by the MIT Generative AI Impact Consortium (MAGIC) and DSO National Laboratories in Singapore. The authors acknowledge the MIT Office of Research Computing and Data for providing high-performance computing resources that have contributed to the research results reported within this paper.



\section*{Impact Statement}

Advances in computational methods for \textit{de novo} structure elucidation can substantially improve the identification of previously unknown compounds, including metabolites that serve as diagnostic biomarkers or illuminate underlying biological processes. These capabilities have broad potential for positive societal impact across healthcare and life sciences, with minimal foreseeable risk.




\bibliography{main}

\bibliographystyle{icml2026}

\newpage
\appendix
\onecolumn
\setlength{\textfloatsep}{8pt plus 2pt minus 2pt}
\setlength{\intextsep}{8pt plus 2pt minus 2pt}
\setlength{\floatsep}{8pt plus 2pt minus 2pt}

\section{Additional Experimental Results}

\subsection{Additional metrics}
\label{sec:addl_metrics}

Following~\citep{bohde2025diffms} we report additional metrics to further contextualize how close \ours predictions are to the true molecule even in cases where the true structure cannot be recovered. In Table~\ref{table:addl_main}, we report the percentage of FRIGID samples that correspond to valid molecules as well as the percentage of Top-1 and Top-10 predictions that are ``close'' and ``meaningful'' structural matches according to the definitions of~\citet{butler2023ms2mol}. Specifically, a molecule is a ``meaningful'' match if it has an RDKit fingerprint Tanimoto similarity $\geq0.4$ compared to the reference structure, and it is a close match if the similarity is $\geq0.675$. These thresholds were selected based on domain-expert annotations, and close structural matches can be considered to be useful in practical settings. As shown in Table~\ref{table:addl_main}, \ours achieves strong performance, including achieving close Top-1 predictions on more than half of NPLIB1 test spectra. We also note that \ours-base achieves similarly strong Top-10 performance and is able to consistently generate meaningful and close matches without any inference-time scaling, however inference-time scaling helps to refine close matches into exact structural matches.

\begin{table*}[!htbp]
\caption{Additional evaluation of molecule validity, and percentage above domain-expert-defined Tanimoto thresholds on NPLIB1 ~\citep{duhrkop2021canopus} and MassSpecGym~\citep{bushuiev2024massspecgymbenchmarkdiscoveryidentification} \textit{de novo} generation datasets. The best performing model for each metric is \textbf{bold} and the second best is \underline{underlined}. $\dag$ indicates our implementations of baseline approaches without public code. Methods are approximately ordered by performance. Definitions of meaningful match (Tanimoto similarity of RDKit fingerprints $\geq0.4$) and close match (Tanimoto similarity of RDKit fingerprints $\geq 0.675$) are taken from \citet{butler2023ms2mol}.
}

\label{table:addl_main}
\begin{center}
\resizebox{\linewidth}{!}{
\begin{tabular}{lccccc}
\toprule
& Overall & \multicolumn{2}{c}{Top-1} & \multicolumn{2}{c}{Top-10} \\
\cmidrule(lr){2-2}
\cmidrule(lr){3-4}
\cmidrule(lr){5-6}
Model & \% Valid $\uparrow$ & \% Meaningful match $\uparrow$ & \% Close match $\uparrow$  & \% Meaningful match $\uparrow$ & \% Close match $\uparrow$ \\
\midrule
\midrule
& & \multicolumn{3}{c}{NPLIB1}  & \\
\midrule
Spec2Mol & 66.5\% & 4.09\% & 0.00\% & 8.05\% & 0.00\% \\
MIST + Neuraldecipher & 81.78\% & 65.81\% & 28.57\% & 78.14\% & 39.68\% \\
MIST + MSNovelist  & \underline{98.60\%} & 32.90\% &  11.78\% & 44.79\% &  19.02\%  \\
MIST + MolForge$^\dag$ & 79.88\% & 68.13\% & 39.19\% & 81.20\% & 52.14\% \\
DiffMS & \textbf{100.0\%} & 77.10\% & 42.82\% & \underline{87.50\%} & 59.03\% \\
\midrule
\ours-base (ours) & \textbf{100.0\%} & \underline{79.49\%} & \underline{49.08\%} & 87.06\% & \underline{60.56\%} \\
\ours (ours) & \textbf{100.00\%} & \textbf{80.59\%} & \textbf{51.89\%} & \textbf{87.91\%} & \textbf{61.29\%} \\
\midrule
& & \multicolumn{3}{c}{MassSpecGym} & \\
\midrule
Spec2Mol & 68.5\% & 4.09\% &  0\% & 27.70\% & 0\% \\
MIST + Neuraldecipher & 91.11\% & 53.75\% & 2.50\% & 63.37\% & 4.27\% \\
MIST + MSNovelist & \underline{98.58\%} & 0.66\% & 0.00\% & 1.92\% & 0.00\% \\
MIST + MolForge$^\dag$ & 73.70\% & 56.39\% & 22.40\% & 79.00\% & 30.84\% \\
DiffMS & \textbf{100.0\%} & 78.56\% & 28.28\% & \textbf{96.67\%} & \underline{51.74\%} \\
FOAM (DiffMS+PubChem) & \textbf{100.0\%} & 78.00\% & \textbf{35.55\%} & \underline{93.66\%} & \textbf{62.94\%} \\
\midrule
\ours-base (ours) & \textbf{100.0\%} & \underline{79.54\%} & 30.20\% & 88.81\% & 40.87\% \\
\ours (ours) & \textbf{100.0\%} & \textbf{80.15\%} & \underline{31.95\%} & 88.45\% & 42.17\% \\
\bottomrule
\end{tabular}
}
\end{center}
\vskip -0.1in
\end{table*}

\subsection{Sensitivity to Refinement Hyperparameters}
\label{appendix:scaling_ablations}

We evaluate sensitivity to the main refinement hyperparameters on NPLIB1. Specifically, we vary the maximum masking probability $p_{\max}$, the number of candidates refined per round ($M$), the number of renoised variants per candidate ($N$), and the hallucination penalty weight $\gamma$. Across the tested settings, performance varies only modestly, suggesting that the refinement procedure is reasonably robust and does not depend on a narrow hyperparameter regime.

\begin{table}[!htbp]
\caption{Sensitivity analysis of \ours refinement hyperparameters on NPLIB1. The procedure is fairly stable across nearby settings of $p_{\max}$, $M$, $N$, and $\gamma$.}
\label{table:hyperparam_sensitivity}
\centering
{\small
\begin{tabular}{cccccccc}
\toprule
$\gamma$ & $p_{\max}$ & $M$ & $N$ & Acc@1 $\uparrow$ & Tanimoto@1 $\uparrow$ & Acc@10 $\uparrow$ & Tanimoto@10 $\uparrow$ \\
\midrule
1 & 0.75 & 12 & 8 & 14.16\% & 0.51 & 25.27\% & 0.58 \\
1 & 0.50 & 12 & 8 & 14.16\% & 0.51 & 23.93\% & 0.58 \\
1 & 0.50 & 6 & 6 & 14.53\% & 0.51 & 26.13\% & 0.58 \\
1 & 0.75 & 16 & 4 & 14.53\% & 0.52 & 25.52\% & 0.58 \\
1 & 0.25 & 12 & 8 & 14.16\% & 0.51 & 24.54\% & 0.58 \\
1 & 1.00 & 12 & 8 & 15.02\% & 0.52 & 25.27\% & 0.58 \\
1 & 0.75 & 8 & 12 & 14.65\% & 0.52 & 25.15\% & 0.58 \\
5 & 0.75 & 12 & 8 & 13.80\% & 0.51 & 24.79\% & 0.58 \\
10 & 0.75 & 12 & 8 & 14.29\% & 0.51 & 24.30\% & 0.58 \\
\bottomrule
\end{tabular}}
\vspace{-2mm}
\end{table}

\subsection{MIST Encoders and MolForge Evaluation}
\label{appendix:molforge_discrepancy}

We use MIST 2M-aug on NPLIB1 and MIST 4M-aug on MassSpecGym for both \ours and our MolForge reimplementation. Each encoder is trained from scratch on the corresponding experimental training split and synthetic spectra, with the training procedure described in Appendix~\ref{appendix:mist_training}.

\textbf{Padding mask.} The original MIST implementation adds a boolean mask to attention logits, assigning padded positions a finite $+1$ offset rather than $-\infty$. Padding remains in the softmax during training and inference.

\textbf{Checkpoint and batch size.} The MIST vDiffMS checkpoint also exhibits strongly batch-dependent performance (Table~\ref{table:molforge_bug}). With the same MolForge decoder, reducing the inference batch size from 24 to 1 lowers fingerprint Tanimoto from 0.812 to 0.457 and Top-1 reconstruction accuracy from 31.75\% to 10.73\%. The unusually high batched performance has not been reproduced from the documented training recipe, so checkpoint reproducibility is a separate issue from the padding-mask error.

With MIST 4M-aug, MolForge and \ours achieve 3.29\% and 9.05\% Top-1 accuracy, respectively. Our MIST + MolForge reimplementation is available at \url{https://github.com/harrylaucngd/MIST-MolForge}.

\begin{table}[!htbp]
\caption{MassSpecGym results with different MIST checkpoints and inference batch sizes. Gray rows report vDiffMS checkpoint diagnostics; $B$ denotes inference batch size.}
\label{table:molforge_bug}
\centering
{\small
\begin{tabular}{lccccc}
\toprule
Model & MIST Tanimoto $\uparrow$ & Top-1 Acc $\uparrow$ & Top-1 Tan $\uparrow$ & Top-10 Acc $\uparrow$ & Top-10 Tan $\uparrow$ \\
\midrule
MIST 4M-aug + MolForge & 0.433 & 3.29\% & 0.31 & 5.65\% & 0.40 \\
\color{gray} MIST vDiffMS + MolForge ($B{=}1$) & \color{gray} 0.457 & \color{gray} 10.73\% & \color{gray} 0.37 & \color{gray} 14.48\% & \color{gray} 0.41 \\
\color{gray} MIST vDiffMS + MolForge ($B{=}24$) & \color{gray} 0.812 & \color{gray} 31.75\% & \color{gray} 0.68 & \color{gray} 40.55\% & \color{gray} 0.74 \\
MIST 4M-aug + \ours & 0.433 & 9.05\% & 0.39 & 12.23\% & 0.43 \\
\bottomrule
\end{tabular}}
\vspace{-2mm}
\end{table}

\subsection{Robustness to Imperfect Intermediate Predictions}
\label{appendix:robustness}

The main paper evaluates the standard setting where the chemical formula is assumed known and the molecular fingerprint is predicted from spectra. Here we test robustness when these intermediate signals are imperfect.

\subsubsection{Robustness to Formula Uncertainty}

To measure robustness to formula annotation errors, we use MIST-CF~\citep{goldman2023mist-cf} to generate the top 5 formula hypotheses for each test entry. To preserve a fair compute budget, we keep the total number of generated molecules fixed and distribute that budget across the 5 candidate formulae. While performance degrades without oracle formulae, \ours remains reasonably robust on NPLIB1.

\begin{table}[!htbp]
\caption{Robustness to unknown chemical formulae. We replace the ground-truth formula with the top 5 MIST-CF hypotheses while keeping the total generation budget fixed. The upper block keeps the MIST fingerprints fixed; the lower block rebuilds them using the known or predicted formula.}
\label{table:formula_unknown}
\centering
{\small
\begin{tabular}{lcccccc}
\toprule
& \multicolumn{3}{c}{Top-1} & \multicolumn{3}{c}{Top-10} \\
\cmidrule(lr){2-4}
\cmidrule(lr){5-7}
Model & Accuracy $\uparrow$ & MCES $\downarrow$ & Tanimoto $\uparrow$ & Accuracy $\uparrow$ & MCES $\downarrow$ & Tanimoto $\uparrow$ \\
\midrule
\multicolumn{7}{c}{NPLIB1} \\
\midrule
\ours-base (fixed FP, known formula) & 13.31\% & 7.58 & 0.50 & 20.15\% & 6.16 & 0.56 \\
\ours-base (fixed FP, predicted formula) & 11.60\% & 9.85 & 0.46 & 16.12\% & 7.46 & 0.53 \\
\ours (fixed FP, known formula) & 14.16\% & 7.16 & 0.51 & 25.27\% & 5.79 & 0.58 \\
\ours (fixed FP, predicted formula) & 12.33\% & 9.39 & 0.48 & 20.76\% & 7.47 & 0.54 \\
\midrule
\multicolumn{7}{c}{NPLIB1: matched raw preprocessing} \\
\midrule
\ours-base (rebuilt FP, known formula) & 13.55\% & 7.51 & 0.50 & 20.15\% & 6.12 & 0.56 \\
\ours-base (rebuilt FP, predicted formula) & 11.97\% & 11.23 & 0.43 & 16.12\% & 9.38 & 0.48 \\
\ours (rebuilt FP, known formula) & 14.29\% & 7.10 & 0.51 & 24.54\% & 5.82 & 0.58 \\
\ours (rebuilt FP, predicted formula) & 11.48\% & 10.90 & 0.44 & 21.00\% & 9.28 & 0.50 \\
\bottomrule
\end{tabular}}
\vspace{-2mm}
\end{table}

\FloatBarrier
\subsubsection{Fingerprint Quality Distributional Analysis}

We also retrospectively bin test spectra by the Tanimoto similarity between the MIST-predicted fingerprint and the ground-truth fingerprint. As expected, reconstruction accuracy and structural similarity improve monotonically with fingerprint quality. Nonetheless, even when the predicted fingerprint has Tanimoto similarity below 0.5, the true molecule can still occasionally be recovered among the Top-10 predictions, indicating that \ours can tolerate nontrivial upstream fingerprint error.

\begin{table}[!htbp]
\caption{Distributional analysis of \ours performance as a function of upstream fingerprint quality. Test spectra are binned by the Tanimoto similarity between the MIST-predicted fingerprint and the ground-truth fingerprint.}
\label{table:fingerprint_bins}
\centering
{\small
\begin{tabular}{lcccc}
\toprule
Metric & [0.0, 0.25) & [0.25, 0.5) & [0.5, 0.75) & [0.75, 1.0] \\
\midrule
\multicolumn{5}{c}{NPLIB1} \\
\midrule
Acc@1 $\uparrow$ & 0.00\% & 0.00\% & 10.71\% & 52.44\% \\
Tanimoto@1 $\uparrow$ & 0.16 & 0.32 & 0.60 & 0.90 \\
Acc@10 $\uparrow$ & 0.00\% & 3.32\% & 28.57\% & 71.95\% \\
Tanimoto@10 $\uparrow$ & 0.19 & 0.39 & 0.71 & 0.93 \\
\midrule
\multicolumn{5}{c}{MassSpecGym} \\
\midrule
Acc@1 $\uparrow$ & 0.00\% & 0.06\% & 8.50\% & 63.05\% \\
Tanimoto@1 $\uparrow$ & 0.15 & 0.30 & 0.55 & 0.87 \\
Acc@10 $\uparrow$ & 0.00\% & 0.35\% & 17.99\% & 69.97\% \\
Tanimoto@10 $\uparrow$ & 0.18 & 0.35 & 0.62 & 0.89 \\
\bottomrule
\end{tabular}}
\vspace{-2mm}
\end{table}

\subsection{Direct Spectra Conditioning Ablation}
\label{appendix:e2e_ablation}

The two-stage spectra-to-fingerprint-to-structure pipeline is the standard design in modern \textit{de novo} elucidation systems, but one might ask whether direct end-to-end spectra conditioning is preferable with a stronger decoder. To test this, we fine-tune our model directly on spectra-conditioned generation and compare against the fingerprint-conditioned backbone. Direct end-to-end conditioning consistently hurts exact reconstruction, as shown on NPLIB1. These results support the use of the intermediate fingerprint as a useful structural prior rather than an unnecessary decomposition.

\begin{table}[!htbp]
\caption{Direct spectra conditioning versus the standard fingerprint-conditioned pipeline. End-to-end conditioning underperforms the two-stage formulation on NPLIB1.}
\label{table:e2e_ablation}
\centering
{\small
\begin{tabular}{lcccccc}
\toprule
& \multicolumn{3}{c}{Top-1} & \multicolumn{3}{c}{Top-10} \\
\cmidrule(lr){2-4}
\cmidrule(lr){5-7}
Model & Accuracy $\uparrow$ & MCES $\downarrow$ & Tanimoto $\uparrow$ & Accuracy $\uparrow$ & MCES $\downarrow$ & Tanimoto $\uparrow$ \\
\midrule
\multicolumn{7}{c}{NPLIB1} \\
\midrule
E2E FT-Base & 5.13\% & 10.27 & 0.34 & 9.40\% & 8.62 & 0.42 \\
E2E FT & 7.69\% & 9.49 & 0.39 & 14.04\% & 7.96 & 0.46 \\
\ours-base (fingerprint-conditioned) & 13.31\% & 7.58 & 0.50 & 20.15\% & 6.16 & 0.56 \\
\ours (fingerprint-conditioned) & 14.16\% & 7.16 & 0.51 & 25.27\% & 5.79 & 0.58 \\

\bottomrule
\end{tabular}}
\vspace{-2mm}
\end{table}

\subsection{Multimodal Spectroscopic Dataset Overlap}
\label{appendix:leakage}

\citet{mismetti2026testtimetunedlanguagemodels} have recently developed a test-time tuning strategy that dynamically adapts a pretrained \textit{de novo} elucidation model to the most relevant experimental spectra at inference time. We have omitted a comparison with their approach because their base model is pretrained on a dataset of simulated spectra~\citep{alberts2024unraveling} which we found contains substantial amounts of both NPLIB1 and MassSpecGym test structures. The full results of this analysis are shown in Table~\ref{table:leakage}, where we find more than 10\% of both the NPLIB1 and MassSpecGym test sets overlap with the training set. Additionally, the models used to build this simulated spectra dataset were likely trained on datasets which contain true NPLIB1 and MassSpecGym test structures. Because of this overlap, it is not possible to fairly compare the method in~\citet{mismetti2026testtimetunedlanguagemodels} to \ours and other baselines. While we did not have the capacity to retrain their model from scratch on cleaned data, we note that their test-time tuning strategy is orthogonal to our inference-time scaling and that future works may benefit from combining both ideas.

\begin{table}[!htbp]
\caption{Analysis of structures in the Multimodal Spectroscopic Dataset~\citep{alberts2024unraveling} also present in the NPLIB1 and MassSpecGym test splits. Two structures are considered to be a match if they have the same connectivity-based (2D) InChIKeys.}
\label{table:leakage}
\vspace{0.1in}
\centering
{\small
\begin{tabular}{l|c|c}
\toprule
\multirow{2}{*}{Dataset} & \multicolumn{2}{c}{Train-Test Overlap} \\
& \# of Structures & \% of Test Set \\
\midrule
NPLIB1 & 122 & 14.90\% \\
MassSpecGym & 1820 & 10.37\% \\
\bottomrule
\end{tabular}}
\vspace{-3mm}
\end{table}

\section{Experimental Details}

\subsection{Model Details}

As detailed in Sec.~\ref{sec:model_arch}, we parametrize $p_\theta(\mathbf{x}_0\mid \mathbf{f},\mathcal{C})$ using a BERT-style masked diffusion language model (MDLM) with an additional cross-attention mechanism added in each transformer layer to inject information from the conditioning sequence. We set the MDLM backbone to a 12-layer Transformer encoder with hidden size $d{=}896$, $H{=}14$ attention heads, and feed-forward dimension $d_{\text{ff}}{=}3072$, using dropout 0.1 throughout. We tokenize SAFE strings using a byte-pair encoding (BPE) tokenizer; the resulting vocabulary size is $|\mathcal{V}|{=}1880$, and we use a maximum token length of 256 (including $\texttt{[BOS]}$/$\texttt{[EOS]}$), padding shorter sequences with $\texttt{[PAD]}$.

\textbf{Formula conditioning.} We encode the precursor formula $\mathcal{C}$ as a fixed-dimensional count vector over a set of 30 common elements, without normalization. For cross-attention conditioning, we represent $\mathcal{C}$ as a fixed-length embedding sequence of length 30 by summing (i) a learnable element embedding, (ii) a learnable count embedding (with counts clipped to a maximum of 200), and (iii) a learned positional embedding. We apply a conditioning mask that zeros out elements with count 0. We do not apply formula conditional dropout during training ($p_{\text{drop},\mathcal{C}}{=}0$).

\textbf{Fingerprint conditioning.} We condition on a 4096-bit radius-2 Morgan fingerprint. We embed the fingerprint as an unordered set of active bits, retaining up to 256 indices following canonical bit order; in our training configuration, the activation threshold is 0 (i.e., all non-zero bits are treated as active). Each active index is mapped to a learnable embedding, and we apply 3 layers of permutation-equivariant self-attention (without positional encodings) to model interactions among active substructures, yielding a fingerprint embedding sequence and a corresponding padding mask. For robustness, we apply fingerprint conditional dropout during training with probability $p_{\text{drop},\mathbf{f}}{=}0.25$.

\textbf{Cross-attention injection.} We inject formula and fingerprint conditions via cross-attention sublayers in every Transformer block (with the same hidden size and number of heads as the backbone). We use a shared cross-attention pathway for the two modalities.

\subsection{Training Details}
\label{appendix:training_details}

\subsubsection{Decoder Training}
\label{appendix:decoder_training}

We pretrain the \ours-base DLM on the SAFE-GPT molecular corpus~\citep{noutahi2023safe}, excluding NPLIB1 and MassSpecGym validation and test structures by connectivity-based (2D) InChIKey. Approximately 17\% of the SAFE strings in the corpus cannot be parsed into valid molecular structures. We retain these samples for pretraining with an all-zero conditioning mask, since their molecular formulae and fingerprints cannot be computed.

We train with AdamW~\citep{loshchilov2019decoupledweightdecayregularization} and cosine learning-rate decay~\citep{loshchilov2017sgdrstochasticgradientdescent}. The largest model is trained on over 1 billion structures. We use the pretrained decoder directly, as fine-tuning on NPLIB1 did not improve end-to-end performance (Table~\ref{table:e2e_ablation}).

The largest model is trained for 1{,}000{,}000 optimization steps with a global batch size of 1024, bfloat16 precision, and gradient clipping at norm 1.0. We use continuous-time training ($T{=}0$), log-linear noise, antithetic time sampling, and sampling $\epsilon{=}10^{-3}$. We maintain an exponential moving average (EMA) of the parameters with decay 0.9999 for evaluation. AdamW uses $\beta_1{=}0.9$, $\beta_2{=}0.999$, $\epsilon{=}10^{-8}$, and zero weight decay. The learning rate increases linearly to $3\times10^{-4}$ over 50{,}000 warmup steps, followed by cosine decay (CosineAnnealingLR) with $T_{\max}{=}400{,}000$ and $\eta_{\min}{=}10^{-8}$. We optimize the MDLM reconstruction objective with global-mean loss aggregation and do not use bracket-SAFE tokens.

We set the auxiliary formula-matching loss weight to zero and prioritize formula-matched candidates at inference time.

\subsubsection{ICEBERG Training}
\label{appendix:iceberg_training}
We train separate ICEBERG models~\citep{wang2025iceberg} on the experimental training structures and spectra of NPLIB1 and MassSpecGym (Table~\ref{tab:iceberg_training_exs}). ICEBERG predicts fragments from a candidate structure and assigns their intensities to produce a simulated MS/MS spectrum.

\begin{table}[ht!]
    \centering
    \small
    \caption{Number of training examples used to train each version of ICEBERG. }
    \begin{tabular}{c|ccc}
    \toprule
        Dataset & Training structures & Training spectra \\
        \midrule
        NPLIB1 & 6,810 & 9,272 \\
        MassSpecGym & 13,543 & 101,573 \\
    \bottomrule
    \end{tabular}
    \label{tab:iceberg_training_exs}
\end{table}

The fragmentation model predicts bond cleavages and is supervised with fragments enumerated and annotated by MAGMa~\citep{Ridder2014-en}. The intensity model takes these fragments and the candidate structure as input. Both models are trained with cross-entropy loss. We further fine-tune the intensity model with a contrastive loss against decoy structures to improve candidate ranking.

\subsubsection{MIST Training}
\label{appendix:mist_training}

\textbf{Synthetic training data.} We train a separate MIST encoder~\citep{goldman2023mist} for each benchmark using experimental training spectra and synthetic spectra from dataset-specific GLACIER simulators~\citep{wang2026glacier}. For NPLIB1, we exclude validation and test connectivity InChIKeys from a 104{,}304{,}929-entry PubChem catalog, then select 2{,}000{,}000 molecules by deterministic hash-priority sampling (seed 20260909). We require at most 100 atoms per molecule and at most 254 occurrences of each element. For each molecule, we simulate two adducts ($[\mathrm{M}+\mathrm{H}]^+$ and $[\mathrm{M}+\mathrm{Na}]^+$), two instruments (Orbitrap and QTOF), and five collision energies (10, 20, 30, 40, and 50\,eV), yielding 40{,}000{,}000 synthetic spectra. We call this encoder MIST 2M-aug.

For MassSpecGym, we use the official MassSpecGym~v1.0 collection of approximately 4 million unlabeled molecules, which excludes molecules within MCES distance 2 of the test set. We remove two molecules with unsupported formula counts, leaving 3{,}922{,}525 molecules. For each, we sample five collision energies without replacement from 0 to 100\,eV in 5\,eV increments and use the same two adducts and instruments. This yields 78{,}450{,}500 synthetic spectrum--molecule pairs, which we mix with 194{,}119 experimental training pairs to train MIST 4M-aug. The MassSpecGym GLACIER simulator is trained on the experimental training split.

\textbf{Architecture and objective.} Both encoders have 72.3M parameters and predict 4096-bit radius-2 Morgan fingerprints from peak-formula annotations, using up to 10 peaks per spectrum. Each uses positional--cosine formula embeddings, pairwise peak features, four peak-attention layers with 16 heads, four fingerprint refinement layers, hidden size 1024, and dropout 0.1. We combine cosine fingerprint loss with soft Tanimoto loss (weight 0.1) and iterative fingerprint supervision (weight 0.4). We augment half of the input spectra with peak dropping (probability 0.1) and multiplicative intensity jitter (log-scale standard deviation 0.2).

\textbf{Optimization.} Each effective batch contains 8{,}192 synthetic pairs and either 900 experimental pairs for NPLIB1 or 300 for MassSpecGym. We train for approximately 50 passes through each synthetic corpus: 244{,}150 optimizer steps on NPLIB1 and 478{,}850 on MassSpecGym. We use AdamW for NPLIB1 and Adam for MassSpecGym, both with peak learning rate $2\times10^{-4}$, weight decay $10^{-6}$, cosine decay, gradient clipping at norm 5, and bfloat16 precision. Warmup lasts 4{,}883 and 9{,}577 steps, and EMA decay is 0.999 and approximately 0.9975, respectively.

\textbf{Checkpoint selection.} We evaluate once per synthetic-data epoch and select the EMA checkpoint with the highest validation fingerprint Tanimoto: step 166{,}022 for NPLIB1 and step 430{,}965 for MassSpecGym. We calibrate binarization thresholds by matching predicted and reference on-bit counts on validation spectra, then fix them at 0.2042523548 for NPLIB1 and 0.1879101470 for MassSpecGym.

Our implementation builds on MIST~\citep{goldman2023mist} with distributed training and synthetic spectrum augmentation. Code, data-preparation scripts, and training configurations are available at \url{https://github.com/coleygroup/MIST}.

\textbf{MIST in arXiv v2 (2604.16648v2).} For MassSpecGym, we selected PubChem molecules whose structures closely resembled those of the test molecules and used ICEBERG to simulate their spectra for MIST training. We excluded exact test structures by connectivity-based (2D) InChIKey, but the selection of augmentation molecules still used test-set structural information in a manner we only later determined to be inappropriate (i.e., biasing the molecules selected for ICEBERG augmentation to be close structural analogs of test structures). Unfortunately, the associated MIST checkpoint was quickly adopted by several subsequent studies, which now confounds comparison of recently reported results. We have attempted to contact all authors of papers that report using this prior MIST checkpoint. 

The revised evaluations use the MIST 2M-aug and 4M-aug encoders trained as described above. Table~\ref{table:mist_version_comparison} compares the results reported in arXiv v2 and v3. FRIGID Top-1 accuracy changes from 25.03\% to 14.16\% on NPLIB1 and from 18.29\% to 9.05\% on MassSpecGym.

\begin{table}[ht!]
\caption{FRIGID results reported in arXiv v2 and v3. Version 3 uses MIST 2M-aug on NPLIB1 and MIST 4M-aug on MassSpecGym. MassSpecGym results are averaged over 17{,}082 test spectra in v2 and all 17{,}556 test spectra in v3.}
\label{table:mist_version_comparison}
\centering
{\small
\begin{tabular}{llcccccc}
\toprule
& & \multicolumn{3}{c}{Top-1} & \multicolumn{3}{c}{Top-10} \\
\cmidrule(lr){3-5}
\cmidrule(lr){6-8}
Model & Version & Accuracy $\uparrow$ & MCES $\downarrow$ & Tanimoto $\uparrow$ & Accuracy $\uparrow$ & MCES $\downarrow$ & Tanimoto $\uparrow$ \\
\midrule
\multicolumn{8}{c}{NPLIB1} \\
\midrule
\ours-base & v2 & 19.80\% & 7.88 & 0.54 & 24.41\% & 6.20 & 0.61 \\
\ours-base & v3 & 13.31\% & 7.58 & 0.50 & 20.15\% & 6.16 & 0.56 \\
\ours & v2 & 25.03\% & 7.10 & 0.58 & 33.37\% & 5.56 & 0.64 \\
\ours & v3 & 14.16\% & 7.16 & 0.51 & 25.27\% & 5.79 & 0.58 \\
\midrule
\multicolumn{8}{c}{MassSpecGym} \\
\midrule
\ours-base & v2 & 16.09\% & 15.02 & 0.41 & 18.19\% & 11.95 & 0.46 \\
\ours-base & v3 & 8.21\% & 13.16 & 0.38 & 9.82\% & 11.65 & 0.43 \\
\ours & v2 & 18.29\% & 13.49 & 0.43 & 22.00\% & 11.65 & 0.47 \\
\ours & v3 & 9.05\% & 12.60 & 0.39 & 12.23\% & 11.41 & 0.43 \\
\bottomrule
\end{tabular}}
\end{table}

\subsection{Inference-time Scaling Details}

Our inference-time scaling procedure implements the cycle-consistent refinement pipeline described in Sec.~\ref{sec:inference_scaling}. For each test spectrum, we obtain a binary fingerprint by thresholding the spectrum-predicted fingerprint probabilities with $\vartheta=0.2042523548$ on NPLIB1 and $\vartheta=0.1879101470$ on MassSpecGym, and treat the molecular formula as known. After each refinement round, we rank the pool of unique candidates by fingerprint Tanimoto similarity to the predicted fingerprint, while prioritizing candidates whose molecular formula matches the target formula; we then report Top-$k$ metrics on the ranked list. Refinement is run for $R$ rounds with a per-round sample budget $B$. In round 1, we generate $B$ candidates from scratch using the conditional DLM. In subsequent rounds, we refine the top $M$ unique candidates from the current pool by constructing $N$ renoised variants per candidate (via masking) and denoising them, yielding up to $N \cdot M$ refined samples; the remaining budget, $B{-}N \cdot M$, is filled with additional from-scratch generations to preserve diversity. By default, we use $B{=}128$, $M{=}12$, and $N{=}8$. For both datasets, we use $R=11$ rounds in total: initial generation followed by 10 refinement rounds, and evaluate intermediate rounds.

We simulate spectra using ICEBERG with collision energies $\{10,20,30,40,50,60,70,80,90\}$ (eV) and adduct $[M{+}H]^+$. We match simulated and experimental peaks within a 20\,ppm tolerance and ignore the precursor peak. For each candidate, we identify a small set of informative simulated peaks (by intensity thresholding and selecting the top few peaks) and mark each peak as \emph{matched} if it lies within the ppm tolerance of any experimental peak; otherwise it is treated as a hallucination. ICEBERG associates each simulated peak with a fragment, enabling attribution of peaks to atoms in the candidate molecule. As described in Sec.~\ref{sec:inference_scaling}, atom-wise consistency scores are computed, then normalized and aggregated into token-wise consistency scores, using rewards and penalties from matched and hallucinated peaks, respectively. Finally, tokens with non-positive scores are independently masked with probability $p_{\max}\cdot |S_{\text{token}}|$, where $p_{\max}{=}0.75$. The resulting partially masked sequence is then re-denoised under the original conditions to propose refined candidates.

We maintain a persistent candidate pool across rounds and deduplicate candidates by connectivity (stereochemistry-agnostic) identifiers to ensure refinement invests compute in genuinely new structures rather than repeated samples. To amortize cost, ICEBERG calls are batched across molecules and collision energies, and simulated spectra are cached for reuse across rounds.

\subsection{Compute Benchmarking Details}

For our inference compute benchmarks (in Sec.~\ref{sec:ablate_speed}),
we measure the latency in seconds/spectrum on a single NVIDIA 6000 Ada GPU after an initial warmup period. Diffusion models, including \ours, DiffMS, and MADGEN, generate many samples and then apply some ranking procedure to determine the final ranked predictions. Specifically, \ours and DiffMS generate 100 samples, and MADGEN generates 50 samples. The reported seconds/spectrum refers to the total time taken to generate all samples (which is different for each model). We omit speed comparisons to Spec2Mol~\cite{litsa2023spec2mol} and MIST + Neuraldecipher~\cite{bohde2025diffms, le2020neuraldecipher} as the inference code reproduced in~\citet{bohde2025diffms} is not publicly available. Similarly, to benchmark the GPU hours required to train each baseline in Sec.~\ref{sec:ablate_train_scaling}, we train each model for one epoch on a single NVIDIA A100 GPU and multiply by the number of epochs used to train each model.  

\subsection{Train Compute Scaling Details}
\label{appendix:train_scaling}

In Table~\ref{table:appdx_training_compute} we provide settings and additional experimental results for each of the models in our training compute scaling study. We train each model for a single epoch with increasingly large numbers of training structures ranging from 3.2M to over 1B. For models with less than 10M training structures, we find that smaller models achieve better empirical performance. We hypothesize that larger models might be appropriate when scaling training to billions of structures.

\begin{table}[H]
\caption{Compute and accuracy evaluations from progressive scaling of the pretraining set. Across the evaluated pretraining sets, Top-1 and Top-10 reconstruction and structural similarity continue to improve with additional molecules observed.}

\label{table:appdx_training_compute}
\centering
{\small
\begin{tabular}{ccccccc}
\toprule
\# Training & \multirow{2}{*}{\# Params} & Training & \multicolumn{2}{c}{Top-1} & \multicolumn{2}{c}{Top-10} \\
\cmidrule(lr){4-5} \cmidrule(lr){6-7}
Structures &  & GPU Hours & Accuracy $\uparrow$ & Tanimoto $\uparrow$ & Accuracy $\uparrow$ & Tanimoto $\uparrow$ \\
\midrule
3.2M & 82.5M & 3.92 & 5.25\% & 0.36 & 9.28\% & 0.43 \\
8M & 82.5M & 9.89 & 8.91\% & 0.42 & 14.04\% & 0.48 \\

25.1M & 182.2M & 54.12 & 10.38\% & 0.46 & 16.85\% & 0.52 \\
49.9M & 182.2M & 107.83 & 12.09\% & 0.48 & 19.66\% & 0.54 \\
112.8M & 182.2M & 243.65 & 13.43\% & 0.50 & 21.86\% & 0.56 \\

1.0B & 247.0M & 2777.77 & 14.16\% & 0.51 & 25.27\% & 0.58 \\
\bottomrule
\end{tabular}
}
\vskip -0.1in
\end{table}

\clearpage
\section{\ours Predictions}
\subsection{NPLIB1 Predictions}
\begin{figure*}[ht!]
    \centering
    \includegraphics[width=\linewidth]{figures/analysis_grid_canopus_successes.png}
    \caption{Examples of selected test spectra from the NPLIB1 dataset where FRIGID makes \textbf{successful} reconstructions. Rows \#1 and \#2: Two examples of immediate success, where the true structure was correctly generated and identified as the Top 1 (as in Row \#1) or Top 2 (as in Row \#2) candidate in Round 1 (where no inference-time correction has yet been applied). The panels here show the top 5 ranked candidates. Rows \#3 and \#4: Examples of an eventual success, where FRIGID fails to create the true molecule in earlier rounds, but eventually corrects its predictions to create the true molecule. Row \#3 shows this eventual success after 3 rounds of inference-time correction; Row \#4 after 5 rounds of correction. } 
    \label{fig:frigid_preds_nplib1_successes}
\end{figure*}

\begin{figure*}[ht!]
    \centering
    \includegraphics[width=\linewidth]{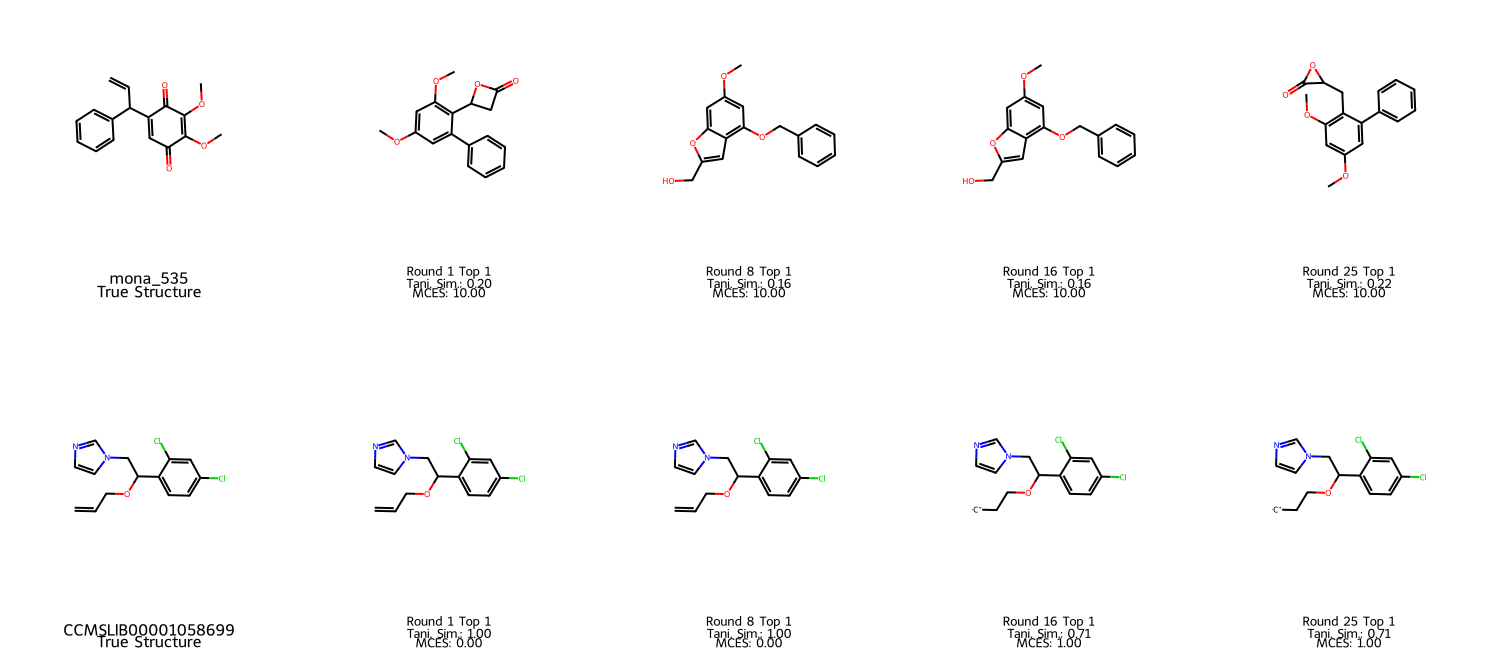}
    \caption{Examples of selected test spectra from the NPLIB1 dataset where FRIGID makes \textbf{errors}. Row \#1: Failure example where FRIGID fails to make or rank the true molecule highly, even after iterative refinement. Row \#2: Failure case where applying inference-time correction worsens the Top 1 ranked structure, from a true structure to a slightly mimicked decoy structure. Note that this structure technically differs in formula and would thus be excluded in a formula filtering step, but we visualize it to highlight a possible failure mode. } 
    \label{fig:frigid_preds_nplib1_fails}
\end{figure*}

\clearpage
\subsection{MassSpecGym Predictions}
\begin{figure*}[ht!]
    \centering
    \includegraphics[width=0.93\linewidth]{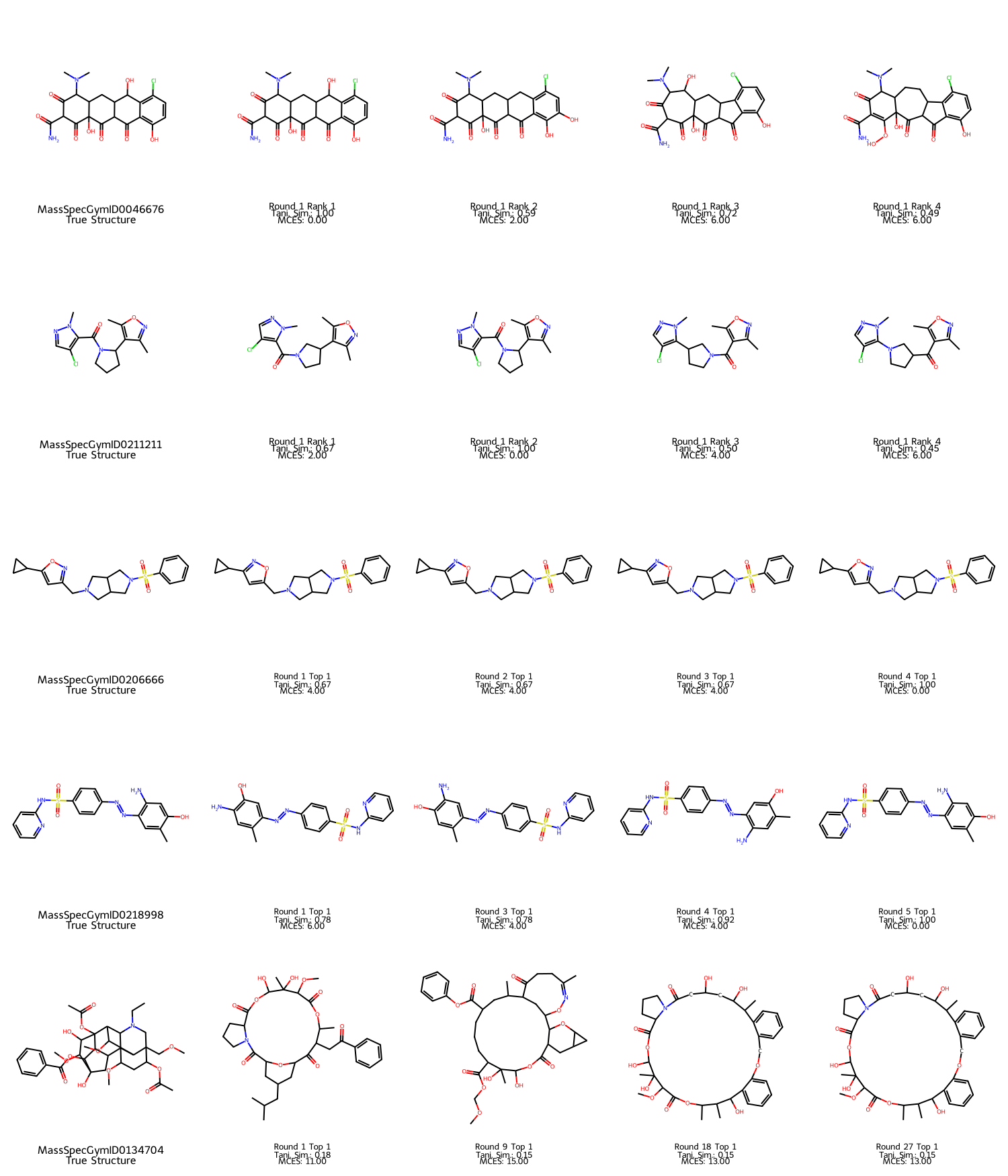}
    \caption{Examples of selected test spectra from the MassSpecGym dataset. Rows \#1 and \#2: Two examples of immediate success, where the true structure was correctly generated and identified as the Top 1 (as in Row \#1) or Top 2 (as in Row \#2) candidate in Round 1 (where no inference-time correction has yet been applied). The panels here show the top 5 ranked candidates. Rows \#3 and \#4: Examples of an eventual success, where FRIGID fails to create the true molecule in earlier rounds, but eventually corrects its predictions to create the true molecule. Row \#3 shows this eventual success after 3 rounds of inference-time correction; Row \#4 after 4 rounds of correction. Row \#4's intermediate molecules show a progressive improvement in structural similarity leading up to the discovery of the true structure. Row \#5: Failure example where FRIGID fails to make or rank the true molecule highly, even after iterative refinement. } 
    \label{fig:frigid_preds_msg}
\end{figure*}


\end{document}